\documentclass[letterpaper, 10 pt, journal, twoside]{ieeetran}  % Comment this line out if you need a4paper

\usepackage[T1]{fontenc}
\usepackage[utf8]{inputenc}
\usepackage{authblk}
\usepackage{graphicx}
\usepackage[spaces,hyphens]{url}
\usepackage{array}
\usepackage{mathtools}
\usepackage{subcaption}
\usepackage{amssymb}
\usepackage{amsmath}
\usepackage{diagbox}
\usepackage{fourier} 
\usepackage{array}
\usepackage{makecell}
\usepackage{hyperref}
\usepackage{stackengine}
\usepackage{multirow}
\usepackage{xcolor, soul}
\sethlcolor{yellow}

\begin{document}

\title{Probabilistic Multi-View Fusion of Active Stereo Depth Maps for Robotic Bin-Picking
}

\author{Jun Yang$^{1}$, Dong Li$^{2}$, and Steven L. Waslander$^{1}$
\thanks{Manuscript received: October, 15, 2020; Revised January, 12, 2021; Accepted March, 06, 2021.}%Use only for final RAL version
\thanks{This paper was recommended for publication by Editor Javier Civera upon evaluation of the Associate Editor and Reviewers' comments. 
This work was supported by Epson Canada Ltd.} %Use only for final RAL version
\thanks{$^{1}$J. Yang and S.L. Waslander are with University of Toronto Institute for Aerospace Studies and Robotics Institute.
        {\tt\footnotesize \{junyang.yang@mail, stevenw@utias\}.utoronto.ca}}
\thanks{$^{2} $D. Li is with Epson Canada Ltd.
        {\tt\footnotesize dong.li@ea.epson.com}}
\thanks{Digital Object Identifier (DOI): see top of this page.}
}
% Use only for final RAL version.

\markboth{IEEE Robotics and Automation Letters. Preprint Version. Accepted March, 2021}
{Yang \MakeLowercase{\textit{et al.}}: Probabilistic Multi-View Fusion of Active Stereo Depth Maps for Robotic Bin-Picking} 

\maketitle
%\thispagestyle{empty}
%\pagestyle{empty}

%%%%%%%%%%%%%%%%%%%%%%%%%%%%%%%%%%%%%%%%%%%%%%%%%%%%%%%%%%%%%%%%%%%%%%%%%%%%%%%%
\begin{abstract}
The reliable fusion of depth maps from multiple viewpoints has become an important problem in many 3D reconstruction pipelines. In this work, we investigate its impact on robotic bin-picking tasks such as 6D object pose estimation. The performance of object pose estimation relies heavily on the quality of depth data. However, due to the prevalence of shiny surfaces and cluttered scenes, industrial grade depth cameras often fail to sense depth or generate unreliable measurements from a single viewpoint. To this end, we propose a novel probabilistic framework for scene reconstruction in robotic bin-picking. Based on active stereo camera data, we first explicitly estimate the uncertainty of depth measurements for mitigating the adverse effects of both noise and outliers. The uncertainty estimates are then incorporated into a probabilistic model for incrementally updating the scene. To extensively evaluate the traditional fusion approach alongside our own approach, we will release a novel representative dataset with multiple views for each bin and curated parts. Over the entire dataset, we demonstrate that our framework outperforms a traditional fusion approach by a 12.8\% reduction in reconstruction error, and 6.1\% improvement in detection rate. The dataset will be available at \url{https://www.trailab.utias.utoronto.ca/robi}.
\end{abstract}


%%%%%%%%%%%%%%%%%%%%%%%%%%%%%%%%%%%%%%%%%%%%%%%%%%%%%%%%%%%%%%%%%%%%%%%%%%%%%%%%

%%%%%%%%%%%%%%%%%%%%%%%%%%%%%%%%%%%%%%%%%%%%%%%%%%%%%%%%%%%%%%%%%%%%%%%%%%%%%%%%%%%%%%%%%%%%%%%%%%%%%%
\section{INTRODUCTION}
\IEEEPARstart{B}{in-picking} is a high value robotic task as it is able to take a tedious, repetitive and dangerous task out of workers' hands. The goal is to have a vision-guided robot to pick up known objects with random poses from a bin. Towards this goal, highly accurate 6D object pose estimation \cite{hinterstoisser2012model,drost2010model,doumanoglou2016recovering,kehl2017ssd,rad2017bb8} is required prior to grasp planning. Due to the requirements of high accuracy and short cycle time in bin-picking, active stereo cameras have been used regularly for this task. Equipped with two cameras and a light projector, its imaging technology simplifies the stereo matching problem and provides reliable depth data. Historically, such cameras are mounted above the bin, and the object pose estimation has been addressed from a static viewpoint. However, from this viewpoint, active stereo cameras often fail to sense complete depths throughout the field of view due to reflective object materials, limited sensor resolutions and occlusions in industrial environments. To overcome these sensing limitations, a multi-view strategy can be employed, but the reliable fusion of these views is a critical step in realizing the benefits of multiple views. 

\begin{figure}[t]
\centering
\begin{subfigure}{.225\textwidth}
  \includegraphics[width=\linewidth]{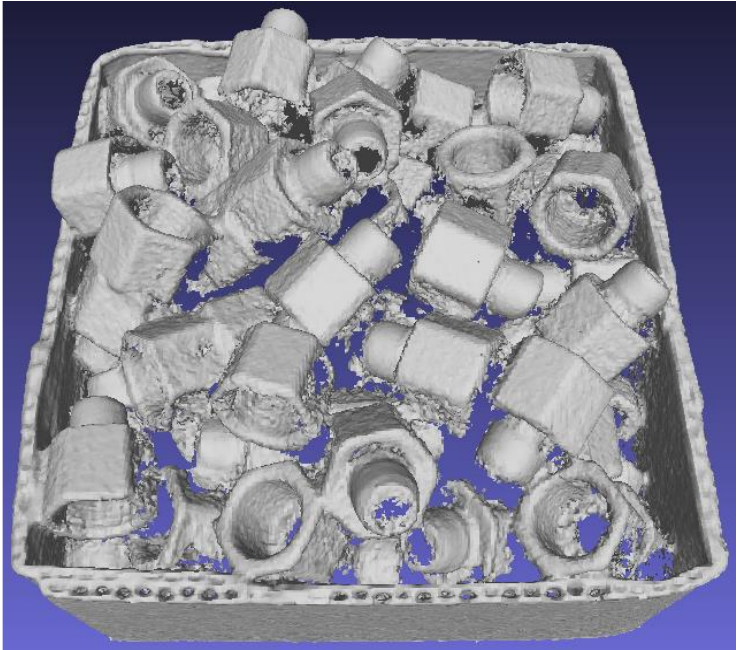}
  \caption{Standard TSDF Fusion}
\end{subfigure}
\begin{subfigure}{.225\textwidth}
  \includegraphics[width=\linewidth]{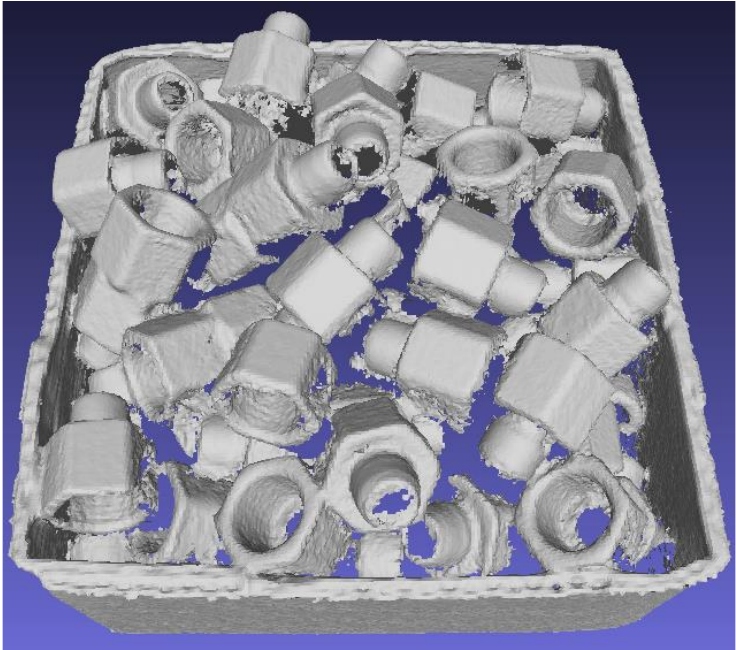}  
  \caption{Ours}
\end{subfigure}
\caption{Standard TSDF Fusion vs. Our fusion approach.}
\label{fig1}
\end{figure}

To efficiently fuse multi-view depth maps, volumetric fusion based on truncated signed distance functions (TSDF) is a commonly used approach \cite{curless1996volumetric,newcombe2011kinectfusion}. Given sequential depth maps with associated camera poses, the TSDF method is able to perform local updates cumulatively via uniformly weighted averaging. The high memory usage for operations on 3D voxel grids can be reduced by voxel hashing \cite{niessner2013real} or octrees \cite{steinbrucker2013large}. All these methods assume that all depth measurements have the same noise level, and are treated equally in TSDF fusion. However, this assumption is usually incorrect since depth uncertainty is typically dependent on the object surface as well as the type of camera sensor and its viewpoint. Moreover, the TSDF fusion does not explicitly handle outliers, therefore depth maps have to be pre-filtered before the fusion step. 

To tackle the limitations of TSDF, a combination of probabilistic modeling and volumetric representation has been proposed in \cite{dong2018psdf}, which fuses measurements using a probabilistic signed distance function (PSDF). This method requires pre-computed depth uncertainty as input for handling noise and outliers. However, the depth uncertainty model is generally difficult to develop since it is determined by various aspects, such as imaging technology, sensor viewpoint as well as scene characteristics. Due to the importance of active stereo camera in robot bin-picking, in this work, we explicitly model depth uncertainties from such sensor at different viewpoints, and incorporate them into a voxel-based probabilistic fusion framework. Our approach is able to perform reliable depth fusion and reconstruct the scene with more details and fewer outliers, as depicted in Figure \ref{fig1}. This will improve 6D object pose estimation performance by providing high quality input depth data.

To demonstrate the advantages of our framework, we present a real-world dataset in bin-picking scenarios. Compared to existing bin-picking related datasets \cite{hinterstoisser2012model, doumanoglou2016recovering, hodan2017t, kleeberger2019large}, our dataset has unique characteristics. It consists of reflective parts and over 30 individual bin instances. We use a robot arm to capture monochrome images and depth maps at multiple viewpoints. For each object, we provide the object model, annotations of 6D object poses, and ground truth depth in real-world.

In summary, we make the following contributions:
\begin{itemize}
    \item A probabilistic framework for scene reconstruction in the bin-picking problem. The framework comprises of a) the explicit estimation of pixel depth uncertainties for active stereo cameras, b) the integration of uncertainty estimates with a probabilistic model for multi-view depth map fusion.
	\item A new multi-view dataset for robotic bin-picking. To the best of our knowledge, this is the first bin-picking dataset with ground truth depth in real-world.
	\item The extensive evaluation of the traditional fusion approach and our method of multi-view depth fusion, as well as its impact on 6D object pose estimation in the bin-picking scenario.
\end{itemize}
%%%%%%%%%%%%%%%%%%%%%%%%%%%%%%%%%%%%%%%%%%%%%%%%%%%%%%%%%%%%%%%%%%%%%%%%%%%%%%%%%%%%%%%%%%%%%%%%%%%%%%

\begin{figure*}[t]
\centering
\includegraphics[width=0.9\textwidth,height=5cm]{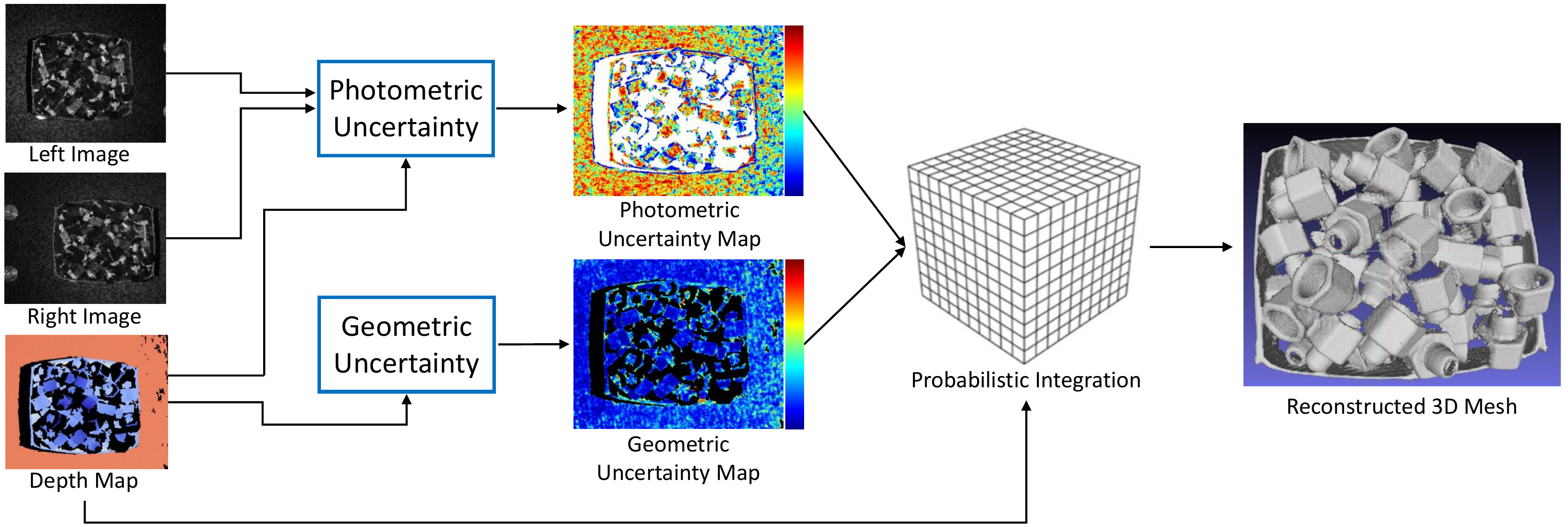}
  \caption{\textbf{Overall system pipeline.} Given depth maps and corresponding active stereo pairs, our system estimates depth uncertainty from both photometric and geometric properties. These uncertainties are then incorporated into a voxel-based probabilistic framework for incrementally updating of the scene.}
  \label{fig2}
\end{figure*}

\section{RELATED WORK}
\noindent\textbf{Object Detection and 6D Pose Estimation.} Accurate 6D object pose estimation is a key step in most robotic bin-picking solutions. In past works, the most representative methods are based on template matching, such as LINEMOD \cite{hinterstoisser2012model}, and its extensions. These methods rely on RGB or RGB-D images as inputs, and are effective in cluttered scene. Point-to-point techniques \cite{drost2010model} represent another category, and are based on depth data only. Many learning-based approaches have been proposed recently to solve 6D object pose estimation problem in an end-to-end manner \cite{kehl2017ssd,rad2017bb8}. The networks are normally trained on synthetic data using 3D CAD models, and tested on real data. Despite the varied pose estimation strategies, the performance of these methods usually relies heavily on the quality of measured depth data.

\noindent\textbf{Multi-View Volumetric Depth Fusion.} When operating limits permit, multi-view fusion is able to provide higher levels of scene completion than single-view methods. In \cite{curless1996volumetric}, Curless and Levoy initially presented an effective method to fuse multi-view depth maps based on TSDF. This method has been later integrated into influential works, such as KinectFusion \cite{newcombe2011kinectfusion}, and further improved with lower memory usage \cite{niessner2013real,steinbrucker2013large}. However, TSDF methods fuse noisy depth maps via uniform weighted averaging in each voxel. Hence, they do not account for the fact that, when observing from different viewpoints, depth measurements may present different levels of noise and outliers. To tackle this limitation, \cite{rozumnyi2019learned} and \cite{weder2020routedfusion} proposed end-to-end learning-based approaches for volumetric depth map fusion. These methods are able to handle sensor noise and outliers, but require a large amount of training data and are relatively prone to over-fitting to a particular sensor or dataset. Based on a model from \cite{vogiatzis2011video}, Wei \textit{et al.} proposed the PSDF Fusion framework \cite{dong2018psdf} for general scene reconstruction. This method requires pre-computed depth uncertainty for incremental Bayesian updating. However, the depth uncertainty is dependent on many scene and sensor characteristics, and is, in general, difficult to acquire.

\noindent\textbf{Uncertainty Estimation for Stereo.} In the context of stereo matching, a wide range of approaches has been proposed for uncertainty estimation. Such estimates were used for outlier removal \cite{hu2012quantitative} and depth map fusion \cite{weder2020routedfusion}. A detailed review of uncertainty estimation for stereo can be found in \cite{hu2012quantitative}. Due to the success of deep learning, some recent works have explored it to learn the uncertainties of disparities intrinsically from data \cite{tosi2018beyond,poggi2016learning}. The ground truth disparities are usually required for the supervised training, which can be difficult to acquire in practice.

\section{METHOD}
Our framework consists of two main parts: (a) uncertainty estimation of active stereo depth maps, (b) volumetric probabilistic integration. In the first part, we use raw disparity maps and corresponding active stereo pairs, and estimate the measurement uncertainties by evaluating photometric uniqueness and geometric consistency. Our uncertainty models are especially useful for determining the pixel-level reliability of depth measurements from active stereo camera sensors. The second part, volumetric probabilistic integration, integrates our uncertainty estimates into a voxel-based probabilistic framework \cite{dong2018psdf}. Based on Bayesian inference, the scene can be updated in an incremental fashion. Figure \ref{fig2} shows an overview of the framework. We assume the camera poses are known, and all computations are performed in a single world coordinate system.

\subsection{Uncertainty from 2D Photometry}
\begin{figure}[t]
\centering
\begin{subfigure}{.225\textwidth}
  \includegraphics[width=\linewidth]{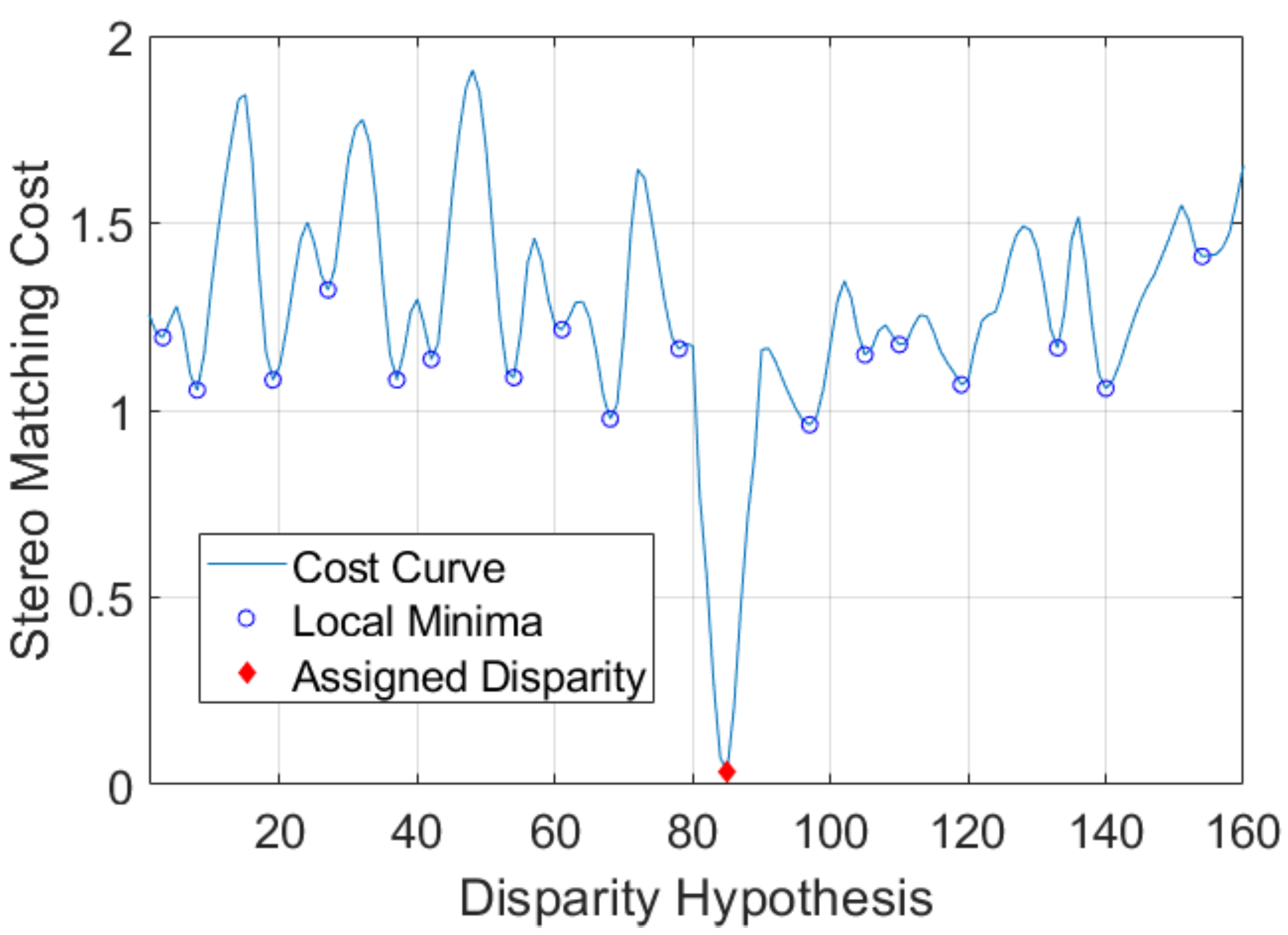}
  \vspace{-1.4\baselineskip}
  \caption{}
\label{figrr1a}  
\end{subfigure}
\begin{subfigure}{.225\textwidth}
  \includegraphics[width=\linewidth]{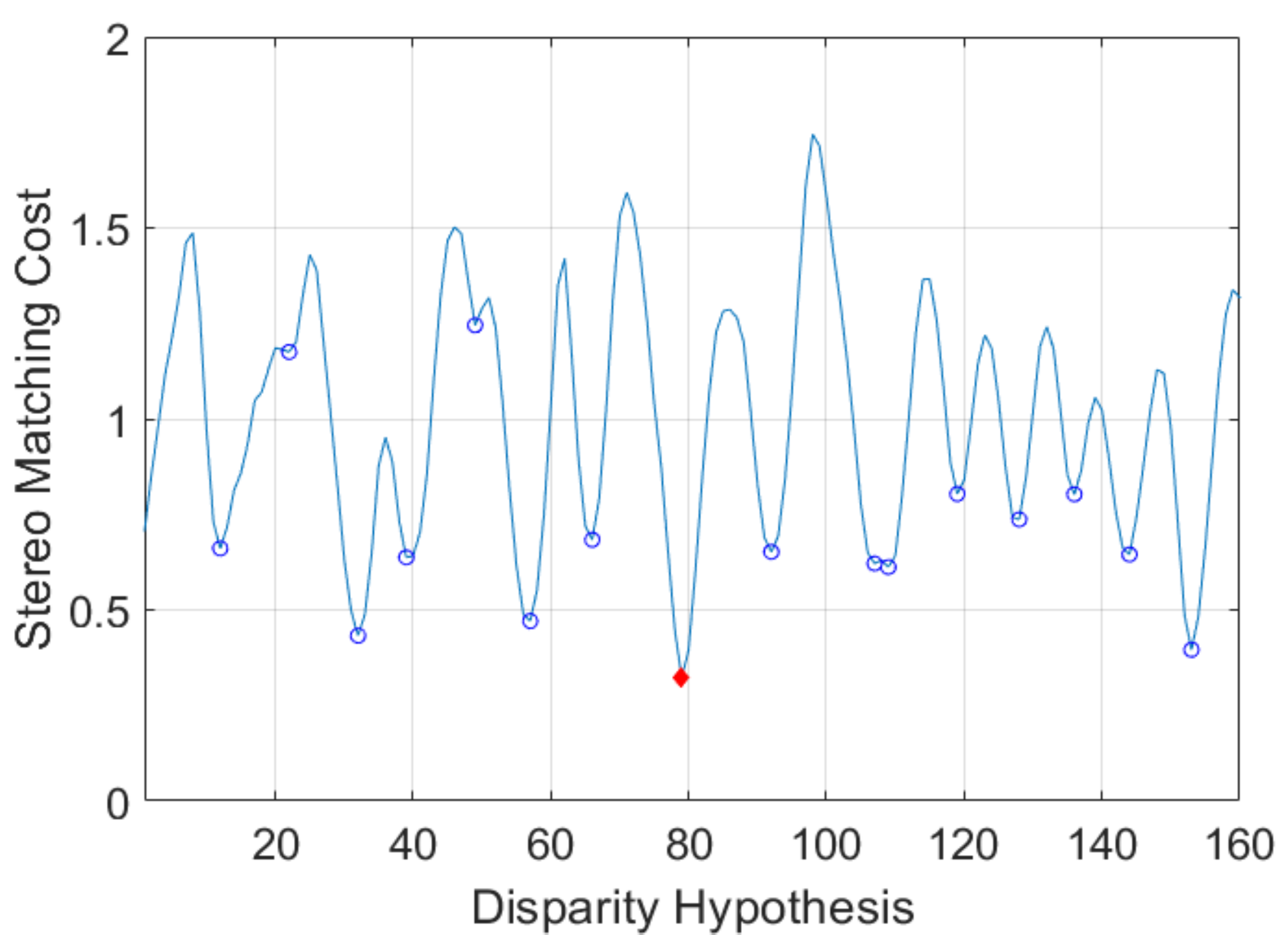}  
  \vspace{-1.4\baselineskip}
  \caption{}
\label{figrr1b}  
\end{subfigure}
\caption{(a) Ideal cost curve. (b) Ambiguous cost curve.}
\label{figrr1}
\end{figure}

To determine photometric uncertainty from active stereo data, we operate on individual pixels by examining their cost curves as a function of the disparity hypothesis. The conventional active stereo camera employs a light projector to create light patterns (e.g., pseudorandom dots) on texture-less regions. This imaging technology simplifies the stereo matching problem by pushing the cost curve to an ideal form. As demonstrated in Figure {\ref{figrr1a}}, the ideal cost curve has a distinct global minimum. However, due to occlusion, shadows and surface reflection, the projected pattern texture is not always visible, which leads to ambiguous cost curves. Figure {\ref{figrr1b}} shows this ambiguity, where multiple local minima with similar cost make localization of the global minimum hard. To determine this uncertainty, we compute a confidence score for each pixel in the raw disparity map. The confidence value indicates the distinctiveness of assigned disparity with adjacent disparity hypotheses.

\begin{figure}[t]
\centering
\begin{subfigure}{0.45\textwidth}
    \includegraphics[width=0.465\linewidth]{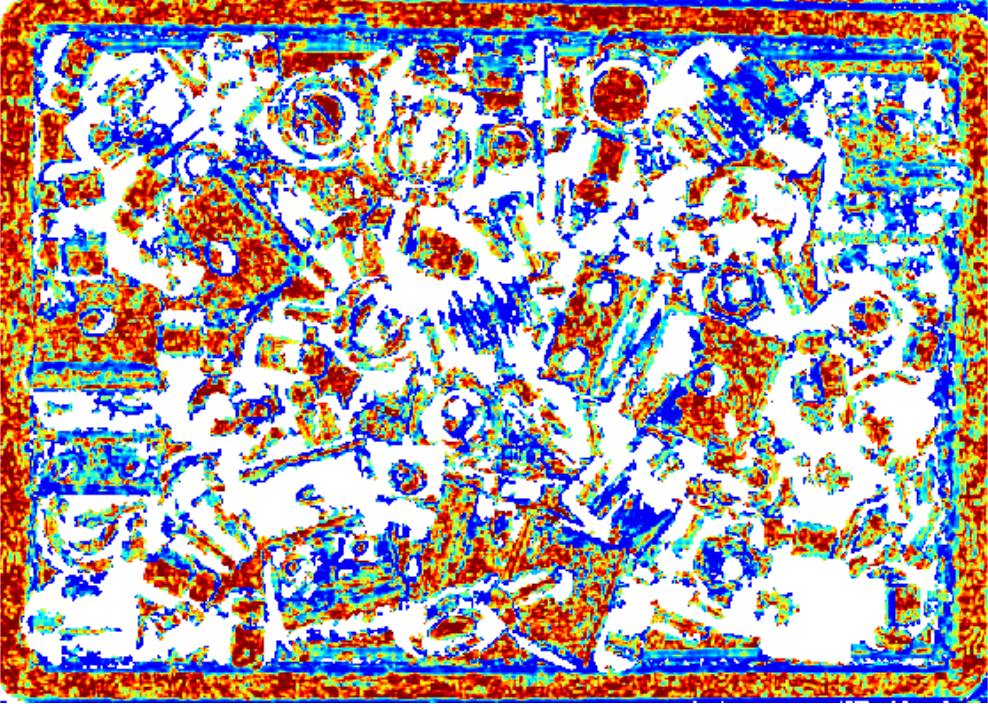}
    \includegraphics[width=0.465\linewidth]{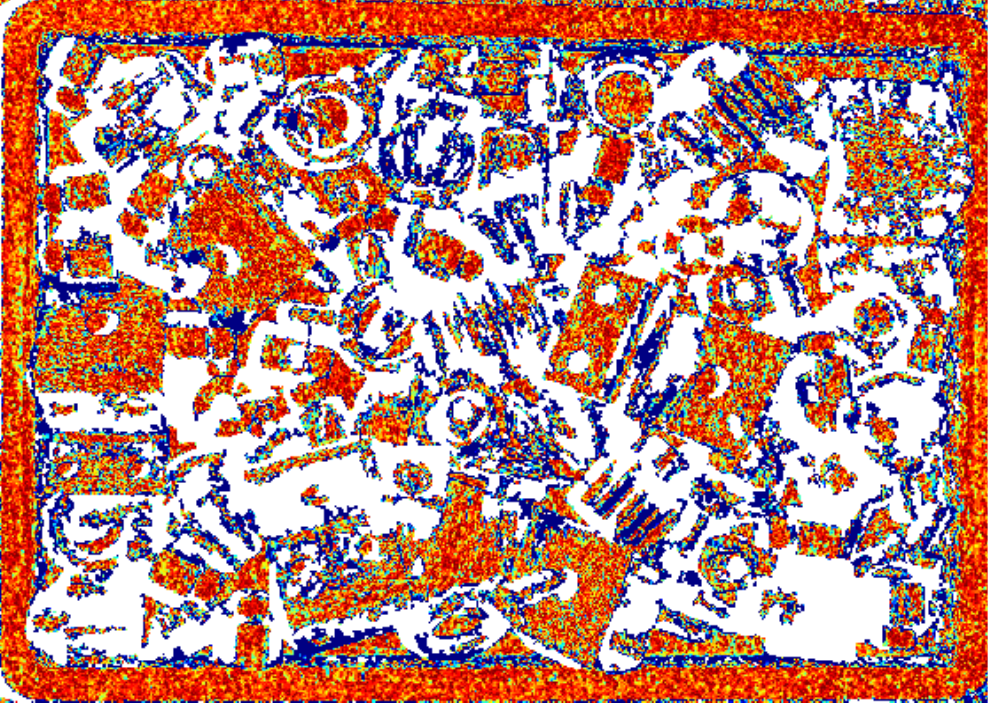}
    \includegraphics[width=0.044\linewidth]{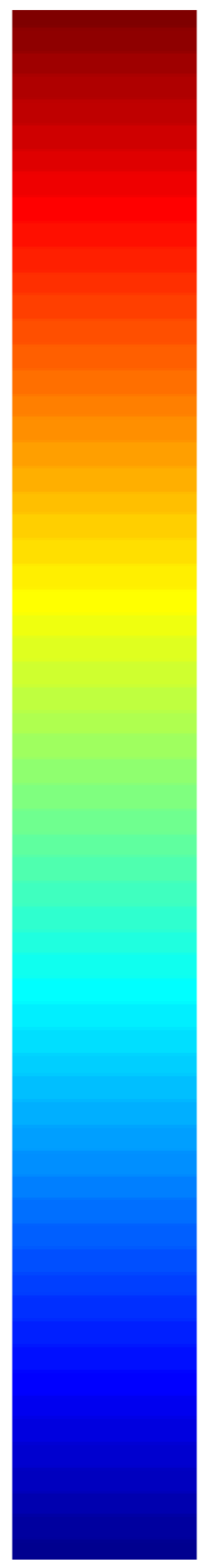}
    \vspace{-1.4\baselineskip}
    \caption{}
\end{subfigure}
\begin{subfigure}{0.45\textwidth}
 \includegraphics[width=\linewidth]{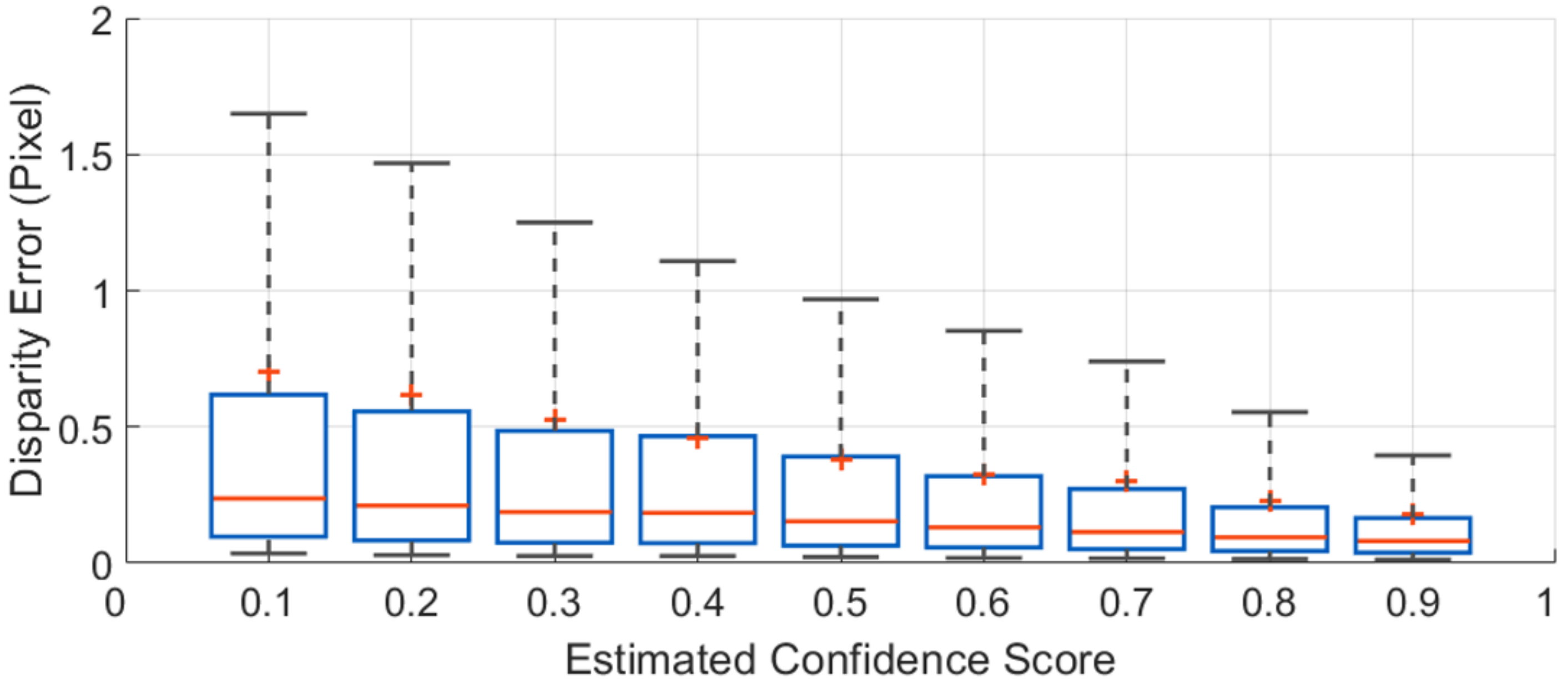}
 \vspace{-1.4\baselineskip}
    \caption{}
\end{subfigure}
\caption{\textbf{Photometric Uncertainty Estimation.} (a) Left: Our confidence measure overlaid on the disparity map. Right: The corresponding error map. Red indicates high confidence (low error) of assigned disparity, and blue low confidence (high error). (b) The correlation between confidence value and disparity error. The paired monochrome images and depth map are shown in Figure \ref{fig6a}.}
\label{fig3}
\end{figure}

Given a stereo pair of rectified, grayscale, left $I_L$ and right $I_R$ pattern projected images, as well as its disparity map $D$, we first assign a cost values $c\left(x,y,d\right)$ to the integer disparity hypothesis $d \in \mathbb{Z}_+$ of pixel $\left(x,y\right) \in I_L$. To reduce illumination sensitivity, we compute the cost value $c: I\times \mathbb{Z}_+ \to \mathbb{C} = [0,2]$ by converting the normalized cross correlation ($NCC$) as $c\left(x,y,d\right) = 1-NCC\left(x,y,d\right)$.  The normalized cross correlation is defined as:
\begin{equation}
NCC\left(x,y,d\right) = \frac{\sum_{i\in W} \left(I_L\left(x_i,y_i\right)-\mu_L\right) \left(I_R\left(x_i-d,y_i\right)-\mu_R\right) }{\sigma_L \sigma_R}
\end{equation}
where $\mu_L, \mu_R \in \mathbb{I}$ denote the means of all pixel intensities within the square window $W$ of the left and right image, respectively, and $\sigma_L$ and $\sigma_R$ are corresponding standard deviations. To compute the confidence score, we choose one of the most promising metrics from \cite{hu2012quantitative}, namely maximum likelihood measure (MLM), and its metric is modified to become:
\begin{equation}
C_{MLM}\left(x,y,d_1\right) = \frac{e^{-\dfrac{c\left(x,y,d_1\right)}{2\sigma^2}}}{\sum_{d'}{e^{-\dfrac{c\left(x,y,d'\right)}{2\sigma^2}}}}
\end{equation}
where parameter $\sigma$ denotes the disparity uncertainty, $c(d_1)$ and $c(d')$ represent the cost value of assigned disparity and local minima, respectively. $MLM$ measures the distinctiveness of assigned integer disparity $d_1$. When margin between $c(d_1)$ and $c(d')$ is larger, it is more likely the assigned disparity is correct. However, the $MLM$ still lacks information about the "strength" of left-right correlation. To account for this, we utilize the original $NCC$ and the final confidence score is computed as:
\begin{equation}
C\left(x,y,d_1\right) = \frac{NCC\left(x,y,d_1\right)+1}{2} \times C_{MLM}\left(x,y,d_1\right)
\end{equation}
For practical reasons, we scale the confidence metric to the interval $[0, 1]$, and the results of our estimated confidence measure are illustrated in Figure \ref{fig3}.

\begin{figure}[t]
\centering
\includegraphics[width=0.45\textwidth]{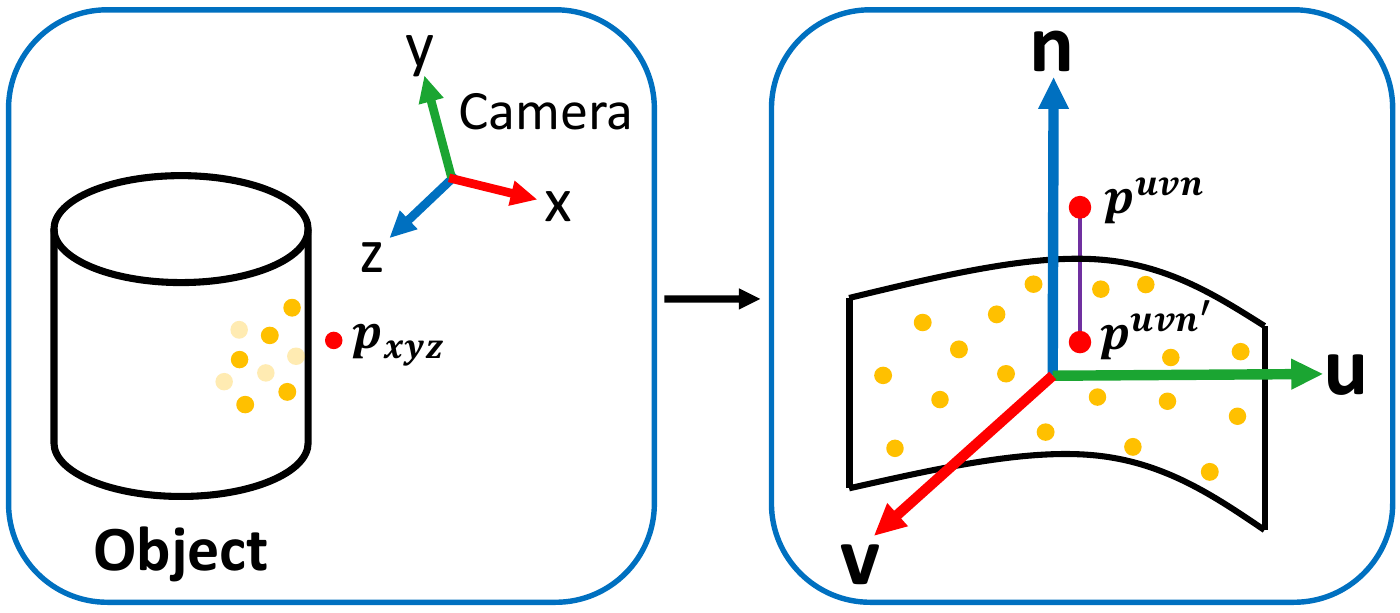}
\caption{The 3D point $\mathbf{p}$ (red dot) and its $N$ nearest neighbors $N(\mathbf{p})$ (yellow dots) are transformed from camera coordinate $xyz$ to local reference frame $uvn$.}
\label{fig4}
\end{figure}

\subsection{Uncertainty from 3D Geometry}
In addition to the photometric uncertainty, the uncertainty of geometry is also calculated in 3D space. We examine consistency of each 3D point $\mathbf{p} \in \mathbb{R}^3$, originating from raw disparity map, $D$, with its local neighbors. For this purpose, we fit a local surface and measure how far $\mathbf{p}$ is to it. We utilize several steps for this computation, which were inspired by data resampling techniques \cite{rusu2008towards}.

For each point, $\mathbf{p}$, from input point cloud, we first project it from camera frame $xyz$ to a local reference frame $uvn$, fitted via Principal Component Analysis (PCA) on its $N$ nearest neighbors, $N(\mathbf{p})$. This is illustrated in Figure \ref{fig4}. Under the coordinate system of $uvn$, a bivariable quadratic function is fit to the heights of the points above the plane:
\begin{equation}
    f(u,v) = a u^2 + b v^2 + c u v + d u + e v + f
\end{equation}
The coefficients $[a, b, c, d, e, f]$ can be computed via a closed-form least squares solution. More details on the computation of polynomial surface can be found in \cite{rusu2008towards}.

\begin{figure}[t]
\centering
\begin{subfigure}{0.45\textwidth}
    \includegraphics[width=0.465\linewidth]{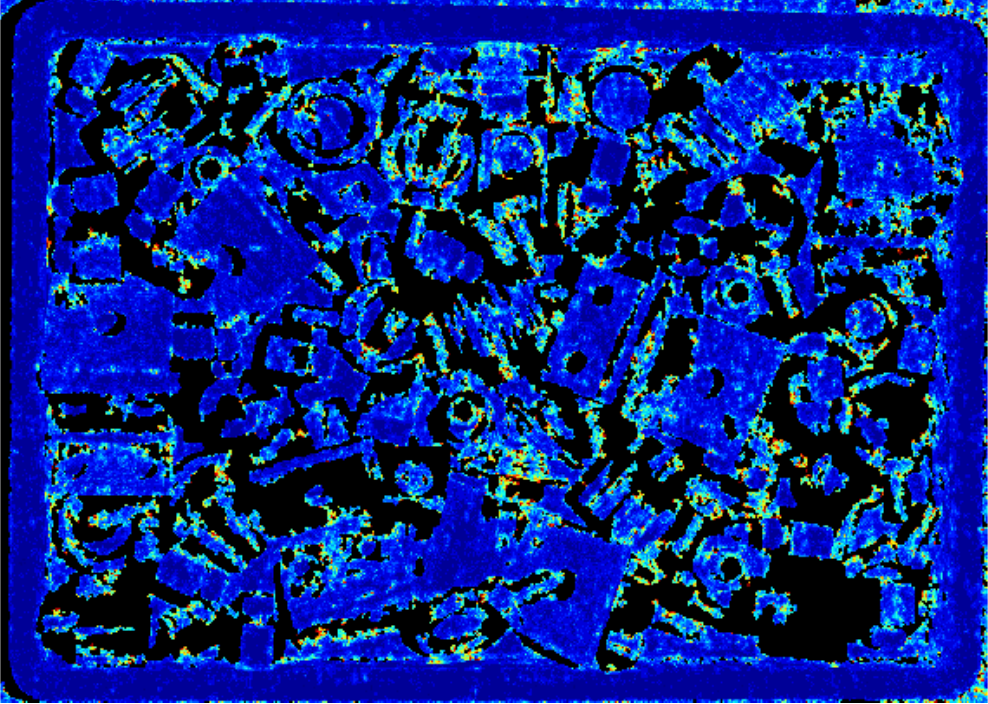}
    \includegraphics[width=0.465\linewidth]{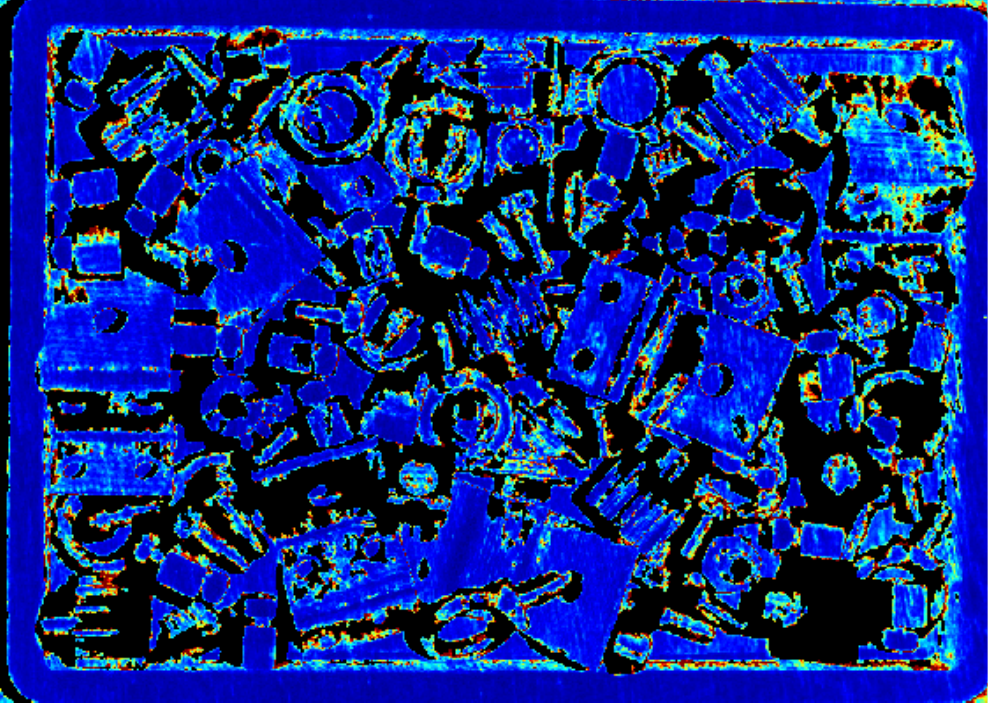}
    \includegraphics[width=0.044\linewidth]{Figs/img_8.pdf}
    \vspace{-1.4\baselineskip}
    \caption{}
\end{subfigure}
\begin{subfigure}{0.45\textwidth}
 \includegraphics[width=\linewidth]{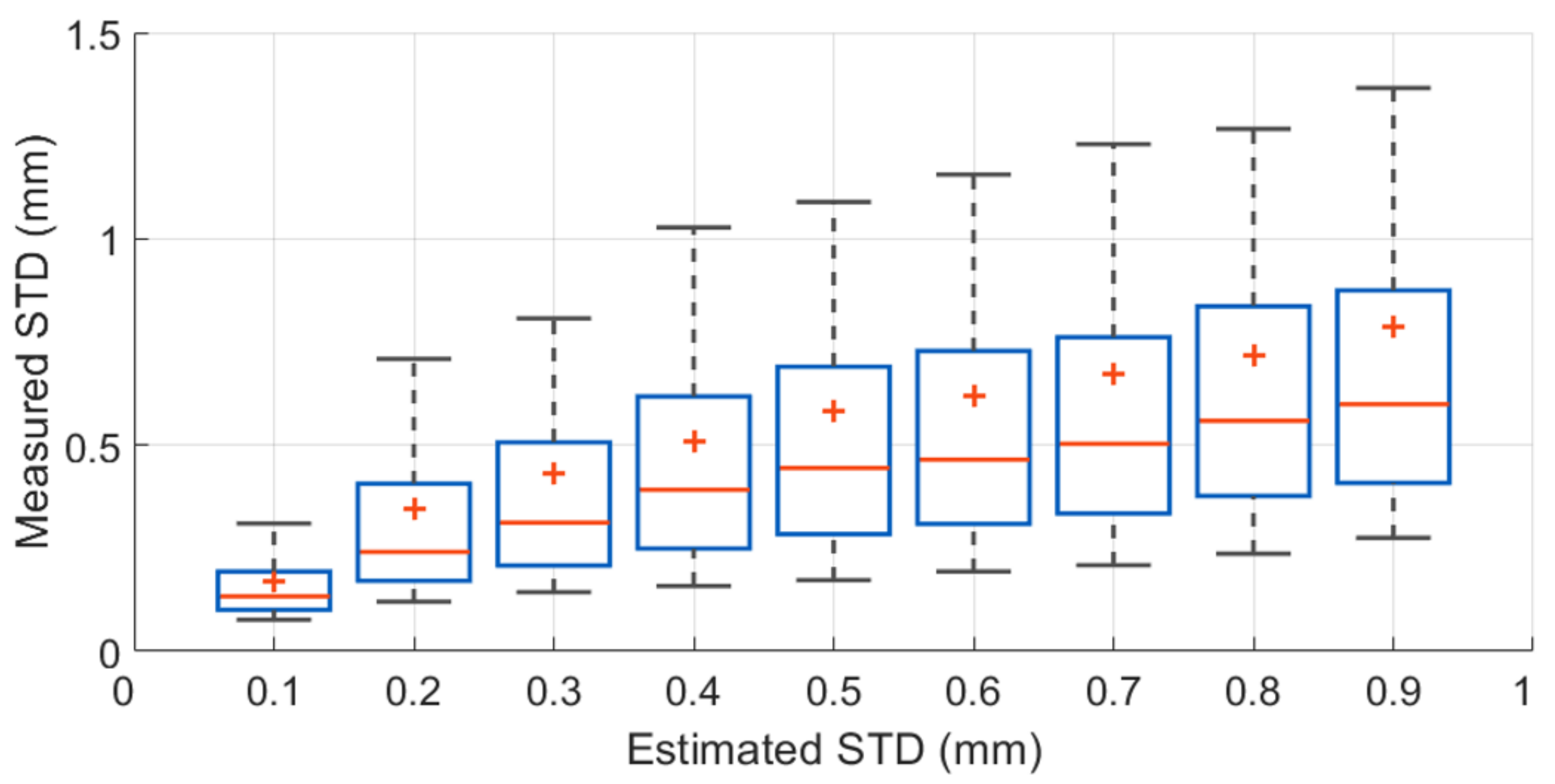}
 \vspace{-1.4\baselineskip}
    \caption{}
\end{subfigure}
\caption{\textbf{Geometric Uncertainty Estimation.} (a) Left: Our estimated STD on depth map. Right: The measured STD from 50 depth maps. Large and small STD of depth measurement are colored in red and blue, respectively. (b) The correlation between estimated and measured STD. The paired monochrome images and depth map are shown in Figure \ref{fig6a}.}
\label{fig5}
\end{figure}

As demonstrated in Figure \ref{fig4}, the signed offset from $\mathbf{p}$ to the local fitted surface is:
\begin{equation}
    \epsilon_{\mathbf{p}} = n_{\mathbf{p}} - n_{\mathbf{p}}^{'}
\end{equation}
where $n_{\mathbf{p}}$ is the height of point $\mathbf{p}^{uvn}$ and $n_{\mathbf{p}}^{'}$ is its recalculated height. The offset for its neighbors $N(\mathbf{p})$ can be computed in the same way, and is represented as $\epsilon_{N(\mathbf{p})}$. With a Gaussian assumption for inlier measurements, we let $\epsilon_{\mathbf{p}}$ be the expectation of the distribution, and treat $\epsilon_{N(\mathbf{p})}$ as its observations. The geometric uncertainty for 3D point $\mathbf{p}$ is finally obtained as the variance of the distribution:
\begin{equation}
\label{equ14}
    \sigma_{\mathbf{p}}^2 = \frac{\sum\epsilon_{N(\mathbf{p})}^2}{N} - \epsilon_{\mathbf{p}}^2
\end{equation}
The results of the estimated and measured geometric uncertainty are illustrated in Figure \ref{fig5}.

\subsection{Probabilistic Volumetric Integration}
Given the depth map $Z_{k} \in \mathbb{R}$ (originating from raw disparity map, with known baseline and focal length) and its camera pose $\mathbf{T}_{k} \in SE(3)$, the classic TSDF fusion \cite{curless1996volumetric,newcombe2011kinectfusion} integrates them to signed distance functions (SDF) $F_k$:
\begin{equation}
    F_{k} = Z_{k} - Z^g
    \label{equ1}
\end{equation}
where $Z^g \in \mathbb{R}$ is a constant value, defined in global frame. To integrate our estimated uncertainties into the global SDF volume, we leverage a per-voxel probabilistic framework, introduced in \cite{dong2018psdf}.

Based on the model proposed in \cite{vogiatzis2011video}, we describe the $k$\textsuperscript{th} SDF observation for a voxel, $F_k$, as a $Gaussian + Uniform$ mixture model distribution. Specifically, an inlier measurement is normally distributed around the true SDF value $\hat{F}$, while an outlier is uniformly distributed in an interval $[F_{min},F_{max}]$:
\begin{equation}
    p\left(F_k|\hat{F},\pi\right) = \pi\textit{N}\left(F_k|\hat{F},\tau_k^2\right)+\left(1-\pi\right)\textit{U}\left(F_k|F_{min},F_{max}\right)
\end{equation}
where $\tau_k^2$ and $\pi$ are the variance and probability of an inlier measurement, respectively. For Bayesian updates, the posterior mixture model can be approximated by the product of a Gaussian and a Beta distribution \cite{vogiatzis2011video} representing the inlier probability:
\begin{equation}
    q\left(\hat{F},\pi|a_k,b_k,\mu_k,\sigma_k^2\right) \overset{\Delta}{=} \textit{Beta}\left(\pi |a_k,b_k\right)\textit{N}\left(\hat{F}|\mu_k,\sigma_k^2\right)
\end{equation}
where $a_n$ and $b_n$ are parameters of the Beta distribution. The update takes the following form:
\begin{equation}
    q\left(\hat{F},\pi|a_k,b_k,\mu_k,\sigma_k^2\right) \approx p\left(F_k|\hat{F},\pi\right)q\left(\hat{F},\pi|a_{k-1},b_{k-1},\mu_{k-1},\sigma_{k-1}^2\right)
    \label{eq:post_dist}
\end{equation}
The true posterior of \eqref{eq:post_dist} by equating first and second order moments for $\hat{F}$ and $\pi$. The updates for $\mu_k$ and $\sigma_k^2$ are derived as:
\begin{equation}
    L_1 = \frac{a_{k-1}}{a_{k-1}+b_{k-1}}\textit{N}\left(F_k|\mu_{k-1},\sigma_{k-1}^2+\tau_k^2\right)
    \label{equ6}
\end{equation}
\begin{equation}
    L_2 = \frac{b_{k-1}}{a_{k-1}+b_{k-1}}\textit{U}\left(F_k|F_{min},F_{max}\right)
    \label{equ7}
\end{equation}
\begin{equation}
    \frac{1}{s^2}=\frac{1}{\sigma_{k-1}^2}+\frac{1}{\tau_k^2},\qquad \frac{m}{s^2}=\frac{\mu_{k-1}}{\sigma_{k-1}^2} + \frac{F_k}{\tau_k^2}
\end{equation}
\begin{equation}
    \mu_k = \frac{L_1}{L_1 + L_2}m + \frac{L_2}{L_1 + L_2}\mu_{k-1}
\end{equation}
\begin{equation}
    \mu_{k}^2 + \sigma_{k}^2 = \frac{L_1}{L_1 + L_2}\left(s^2+m^2\right)+\frac{L_2}{L_1 + L_2}\left(\sigma_{k-1}^2+\mu_{k-1}^2\right)
    \label{equ10}
\end{equation}
The computation of $a_k$ and $b_k$ are the same as \cite{vogiatzis2011video} and are excluded here for compactness.

Due to the linear form in (\ref{equ1}), we use our geometric uncertainty directly as the approximation of the variance of the inlier SDF: $\tau_k \approx \sigma_\mathbf{p}$. Note $\sigma_\mathbf{p}$ is computed per-frame for updating Equations (\ref{equ6}) - (\ref{equ10}). To fully take advantage of estimated uncertainties, a per-frame inlier probability $\pi_k$ is computed for $F_k$ in \cite{dong2018psdf}, to replace the expectation of $\textit{Beta}(\pi |a_{k-1},b_{k-1})$ in Equation (\ref{equ6}) and (\ref{equ7}). Inspired by this, we employ the estimated photometric uncertainty $C$ for evaluating inlier probability $\pi_k$. For this purpose, a mapping from $C$ to the actual inlier probability is required. We leverage a training-based approach from \cite{pfeiffer2013exploiting}, and construct the mapping as:
\begin{equation}
    \pi_k = p\left(\mathbf{i}|C\right)=\frac{p\left(C|\mathbf{i}\right) \cdot p\left(\mathbf{i}\right)}{p\left(C|\mathbf{i}\right) \cdot p\left(\mathbf{i}\right) + p\left(C|\mathbf{o}\right) \cdot \left(1-p\left(\mathbf{i}\right)\right)}
\end{equation}
where $\mathbf{i}$ and $\mathbf{o}$ represent inlier and outlier, respectively. By using the labeled training data, we are able to extract $p\left(C|\mathbf{i}\right)$, $p\left(C|\mathbf{o}\right)$ as well as $p\left(\mathbf{i}\right)$. The computed $\pi_k$ is finally used to replace the simple $\frac{a_{k-1}}{a_{k-1}+b_{k-1}}$, and $\pi$ is still parameterized by $a$ and $b$.

\begin{figure}[t]
\centering
\begin{subfigure}{.16\textwidth}
  \includegraphics[width=\linewidth]{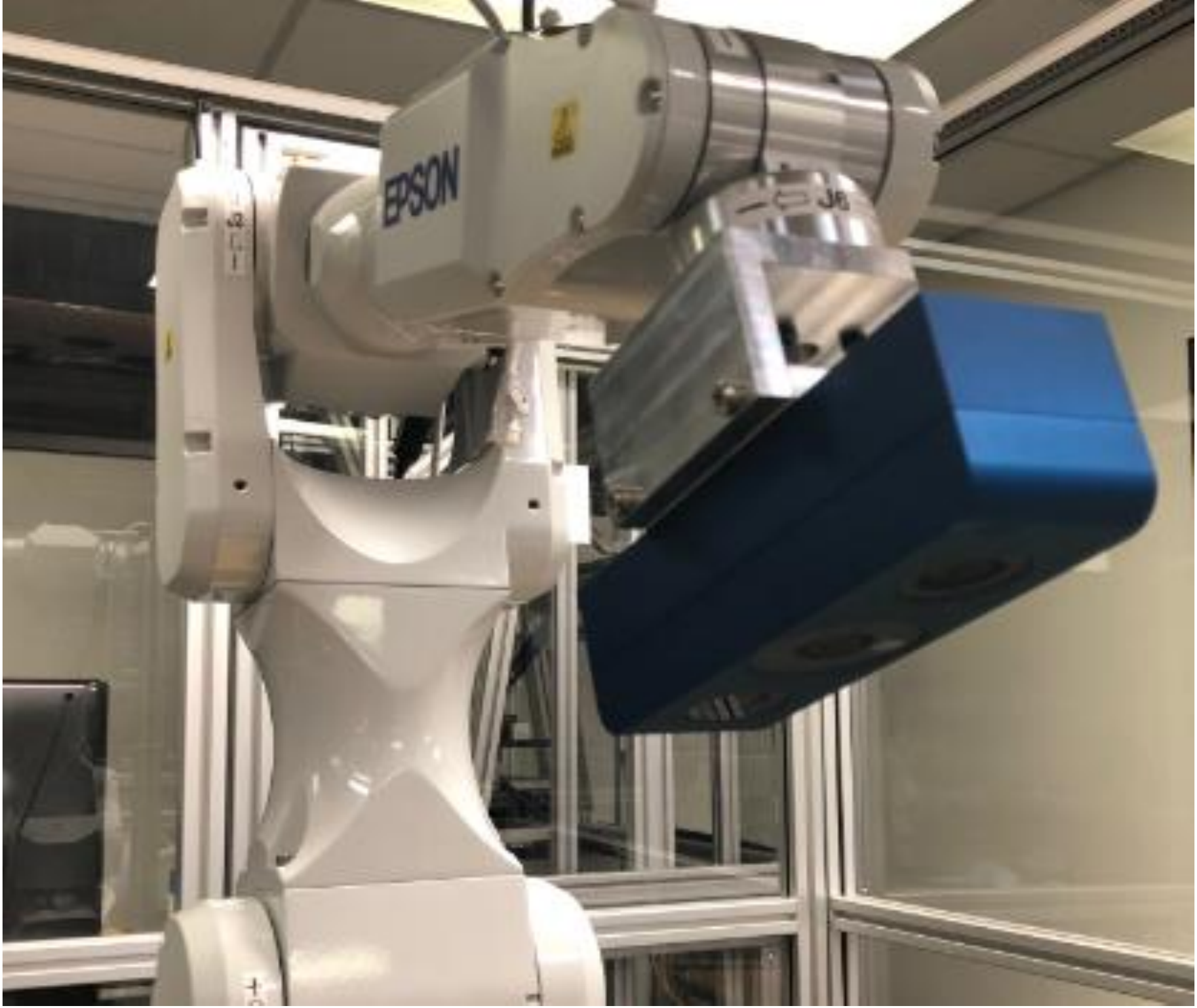}
  \vspace{-1.4\baselineskip}
  \caption{}
\label{fig7a}
\end{subfigure}
\begin{subfigure}{.285\textwidth}
  \includegraphics[width=\linewidth]{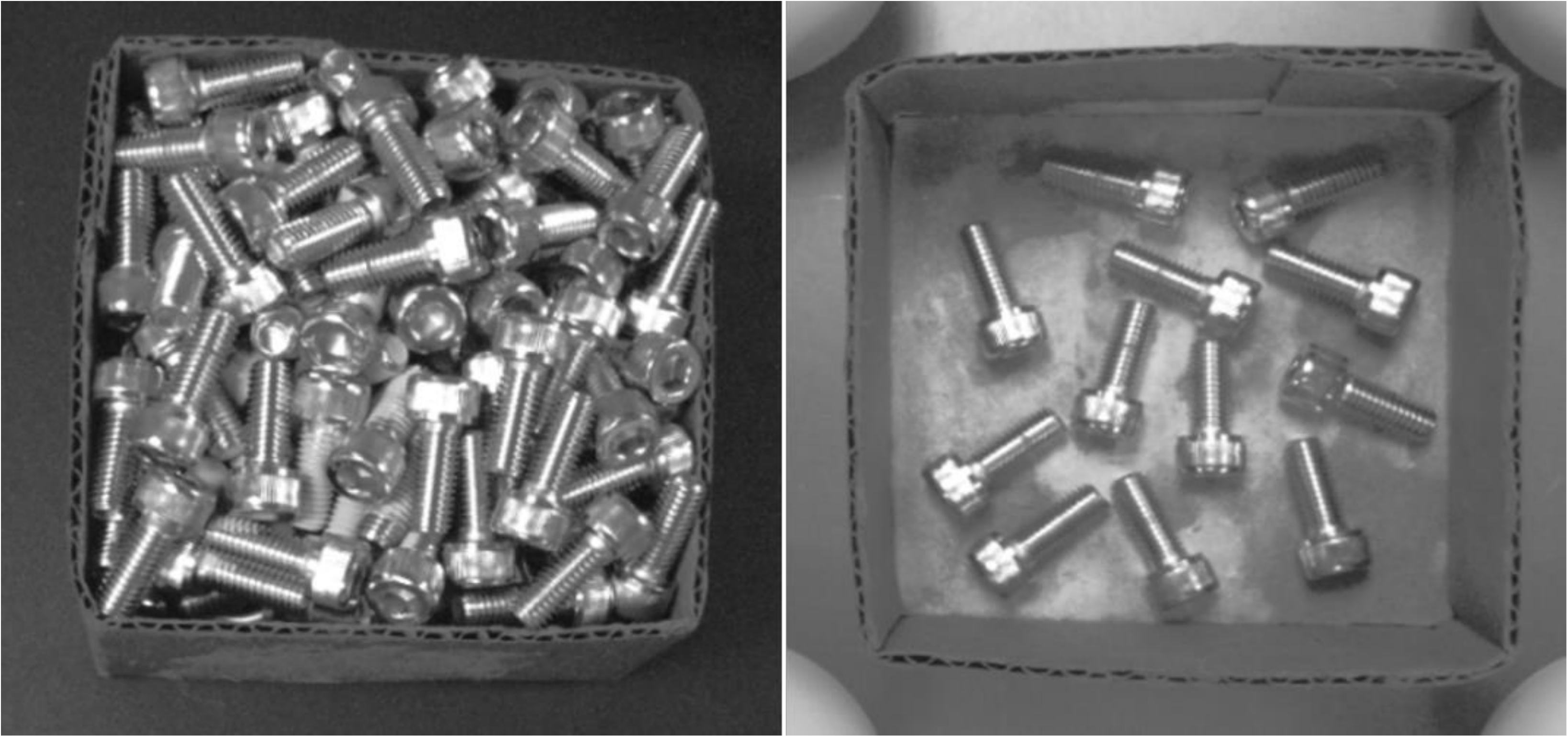}  
  \vspace{-1.4\baselineskip}
  \caption{}
  \label{fig7b}
\end{subfigure}
\begin{subfigure}{.45\textwidth}
  \includegraphics[width=\linewidth]{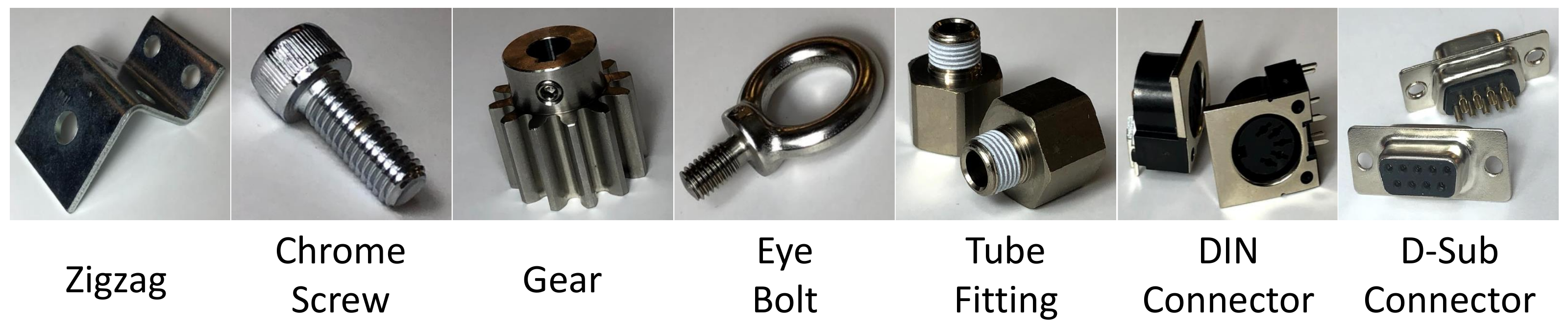}  
  \vspace{-1.4\baselineskip}
  \caption{}
  \label{fig7c}
\end{subfigure}
\caption{\textbf{Overview of ROBI dataset.} (a) we mount an Ensenso N35 camera on an EPSON C4L robot arm. (b) Two bin-picking scenarios. Left: Full-bin scene. Right: Low-bin scene. (c) An overview of all objects of the dataset.}
\label{fig7}
\end{figure}

\subsection{Surface Extraction}
\label{sec3D}
Given SDF values, we use the Marching Cubes algorithm \cite{lorensen1987marching} to extract connected surfaces. Having sufficient photometric and geometric information within voxels, we check their convergence and determine zero-crossing points when the following conditions are satisfied:
\begin{eqnarray}
 \mu^{v_1} \quad\cdot & \mu^{v_2}& < 0, \\
\sigma^{v_1} < \sigma_{thr} & \mbox{and} &  \sigma^{v_2} < \sigma_{thr},\\
\frac{a^{v_1}}{a^{v_1}+b^{v_1}} > \pi_{thr} & \mbox{and} & \frac{a^{v_2}}{a^{v_2}+b^{v_2}} > \pi_{thr},
\end{eqnarray}
where $v_1$ and $v_2$ are adjacent voxels. The threshold $\sigma_{thr}$ and $\pi_{thr}$  control the estimation convergence and can be set according to the accuracy requirements of robotic bin-picking. This operation will reject outliers while preserving distinct features of the scene, which leads to higher quality data for later object pose estimation.

\section{REFLECTIVE OBJECTS IN BINS DATASET}
\label{sec4}
To demonstrate the advantages of our framework, we provide ROBI (Reflective Objects in BIns), a multi-view dataset consisting of reflective objects in bin-picking scenes. For each scene and viewpoint, monochrome images, depth maps and annotations of 6D object poses are included. In addition, we provide ground truth depth maps captured by an active stereo camera with parts coated in anti-reflective scanning spray. We will release a public version of our dataset upon publication of this work, which can be used for scene reconstruction, 6D object pose estimation and depth completion tasks.

\subsection{Sensor Setup}
We captured the raw multi-view sensor data with an Ensenso N35 active stereo camera. The camera is equipped with a short working distance optical lens, and has the working distance from $240\,mm$ to $520\,mm$. We captured images with following two configurations:
\begin{itemize}
\item \textbf{Depth configuration.} We turn camera projector on and use a low exposure time to eliminate the impact of ambient light. The raw disparity maps and pattern projected stereo pairs are collected.
\item \textbf{Monochrome configuration.} The camera projector is turned off for this configuration. We use a high exposure time to obtain optimal contrast for objects. Only stereo pair data is saved with this configuration.
\end{itemize}

\begin{figure}
\centering
    \begin{subfigure}{0.45\textwidth}
            \includegraphics[width=0.32\linewidth]{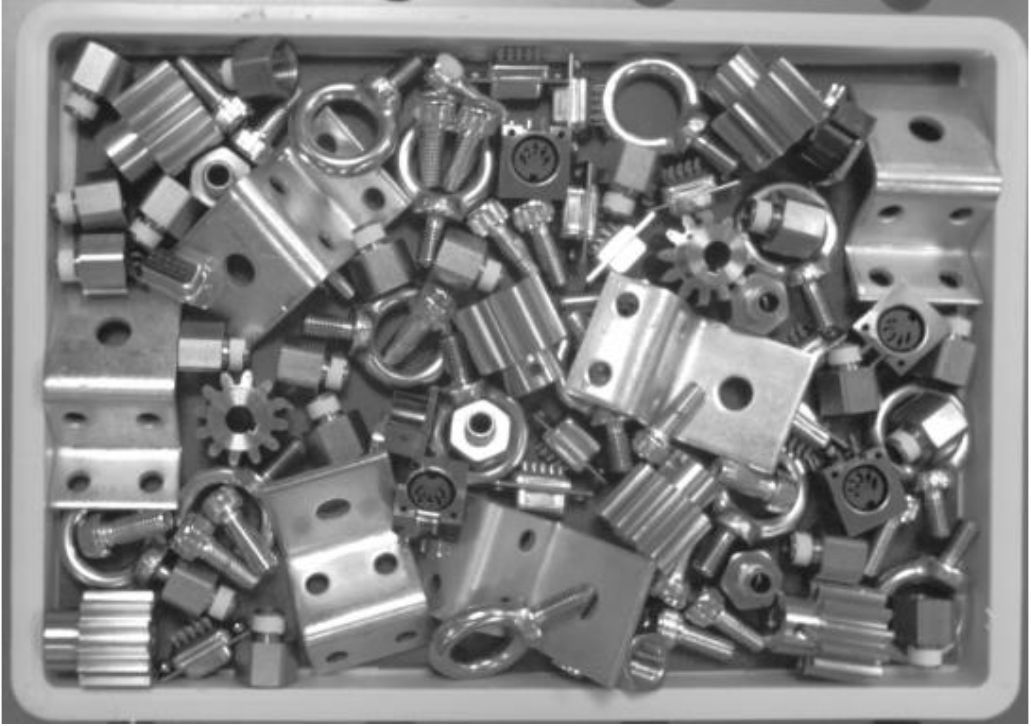}
            \includegraphics[width=0.32\linewidth]{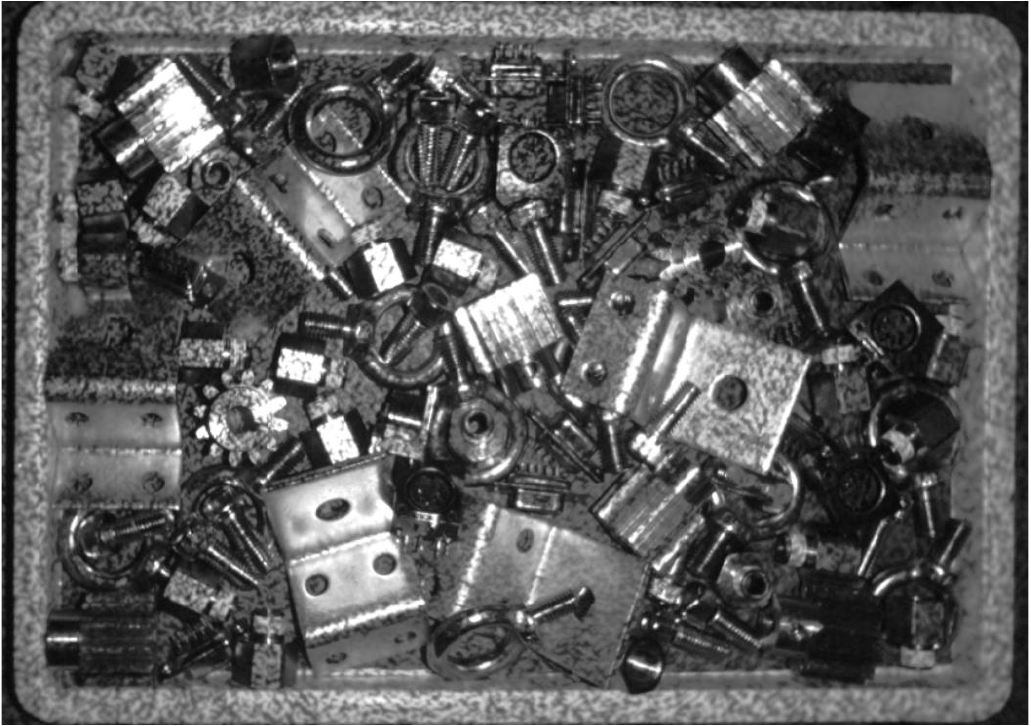}
            \includegraphics[width=0.32\linewidth]{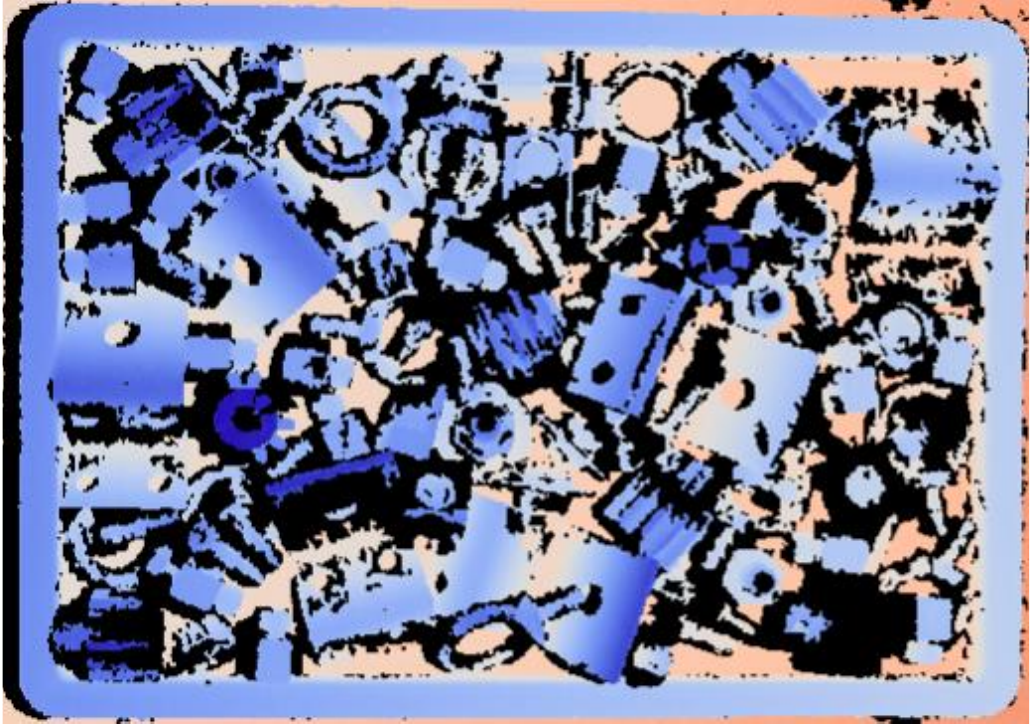}
            \vspace{-0.3\baselineskip}
            \caption{Objects without scanning spray}
    \label{fig6a}                
    \end{subfigure}
    \begin{subfigure}{0.45\textwidth}
            \includegraphics[width=0.32\linewidth]{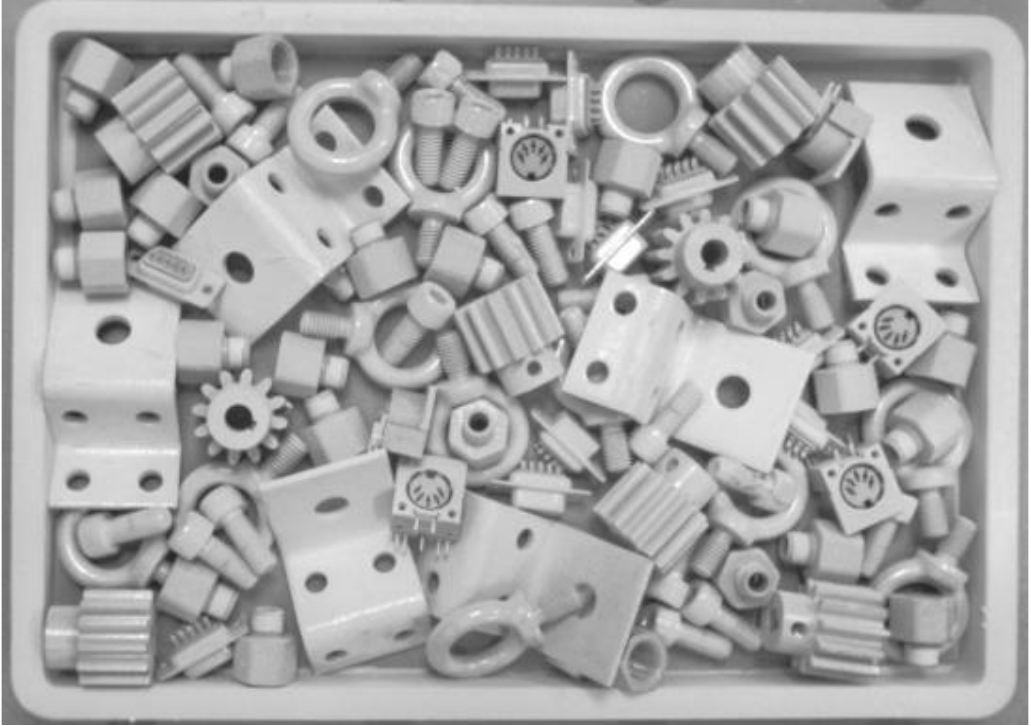}
            \includegraphics[width=0.32\linewidth]{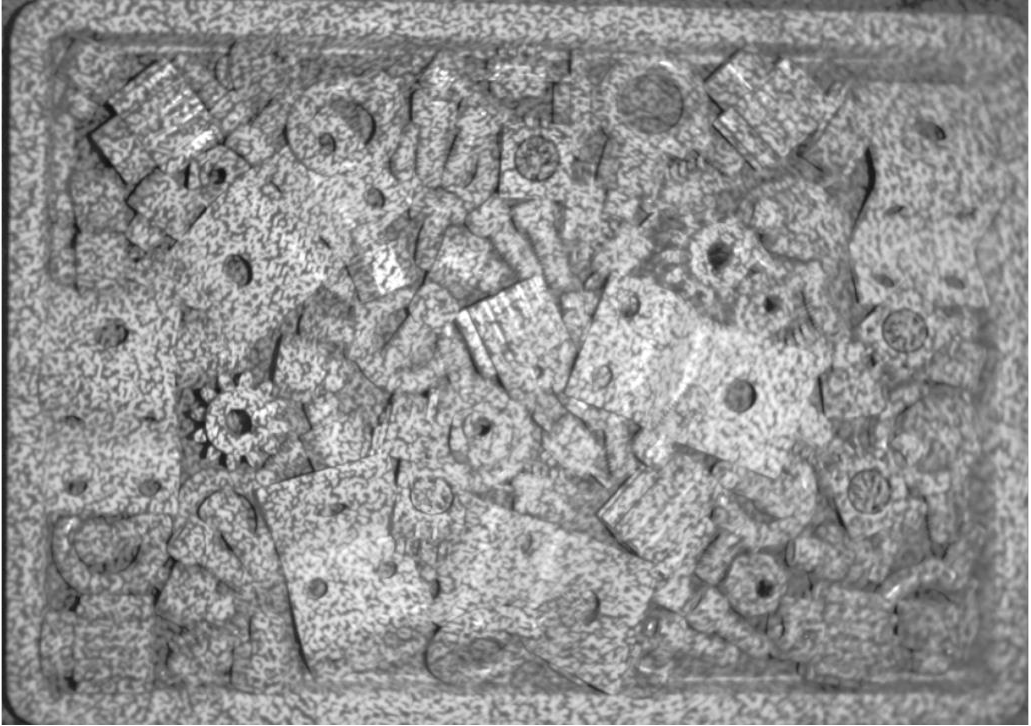}
            \includegraphics[width=0.32\linewidth]{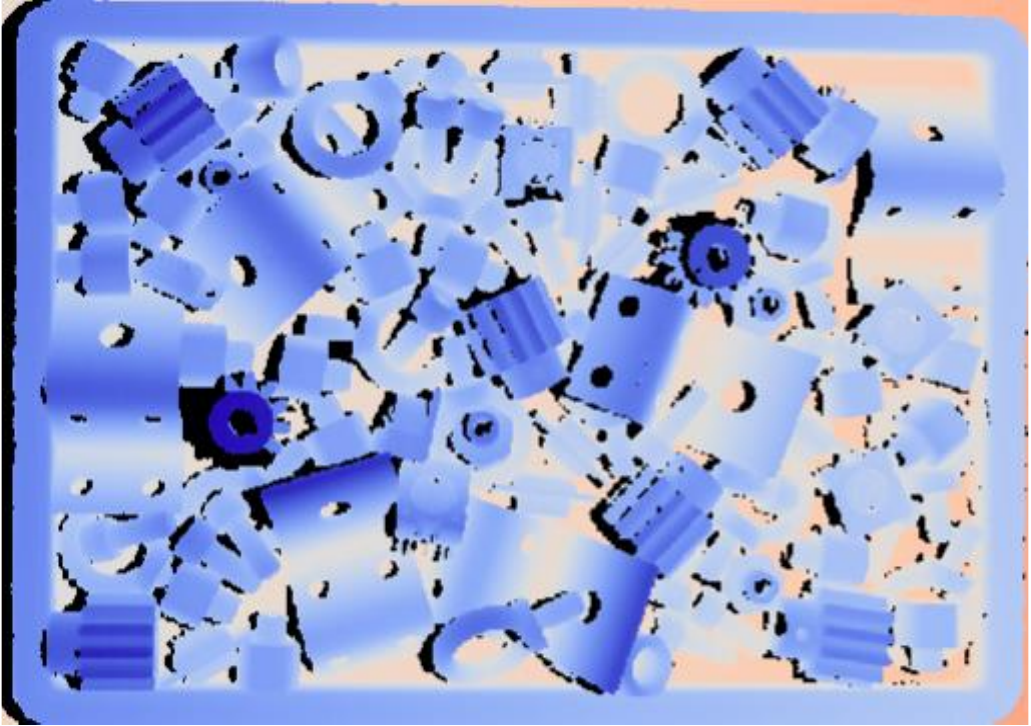}
            \vspace{-0.3\baselineskip}
            \caption{Objects with scanning spray}
    \end{subfigure}
\caption{\textbf{Illustration of using scanning spray on objects.} \textbf{Left}: Left stereo image from monochrome configuration. \textbf{Middle}: Left stereo image from depth configuration. \textbf{Right}: Depth map from depth configuration.}
\label{fig6}    
\end{figure}

\subsection{Data Capture Pipeline}
The camera is mounted to an EPSON 6-Axis C4L robot arm, illustrated in Figure \ref{fig7a}. We program the robot arm to move in a trajectory that traverses a spherical dome and approaches the bin. The robot end-effector stays pointed towards the center of the workstation. The initial camera extrinsics are obtained by leveraging robot end-effector poses and an offline "eye-in-hand" calibration \cite{daniilidis1999hand}. To further refine camera poses, we apply the iterative closest point (ICP) algorithm on calibration spheres, which are placed around the bin. The average closest-point residual error was successfully reduced from $0.33\,mm$ to $0.26\,mm$.

We use seven challenging industrial objects, with different levels of reflectivity. An overview of all objects is depicted in Figure \ref{fig7c}. To demonstrate the bin-picking problem in realistic conditions, we separate our data capture into 2 scenarios: (a) full-bin: multiple objects are stacked on a bin with severe occlusions and clutter, (b) low-bin: a small number of objects are used, with spatial separation between parts. These two scenarios are shown in Figure \ref{fig7b}.

For each object, we capture 5 bin scenarios (3 for full-bin and 2 for low-bin). We sample a total of 88 views for full-bin scenarios, from approximately $45^{\circ}$ to $90^{\circ}$ of sphere elevation, with distances from $400\,mm$ to $520\,mm$. Due to the occlusion of bin wall, in low-bin data capture, the sphere elevation is limited to the range of $[65^{\circ}, 90^{\circ}]$, and there are 42 views in total.

\begin{figure}[t]
\centering
\begin{subfigure}{0.45\textwidth}
  \includegraphics[width=\linewidth]{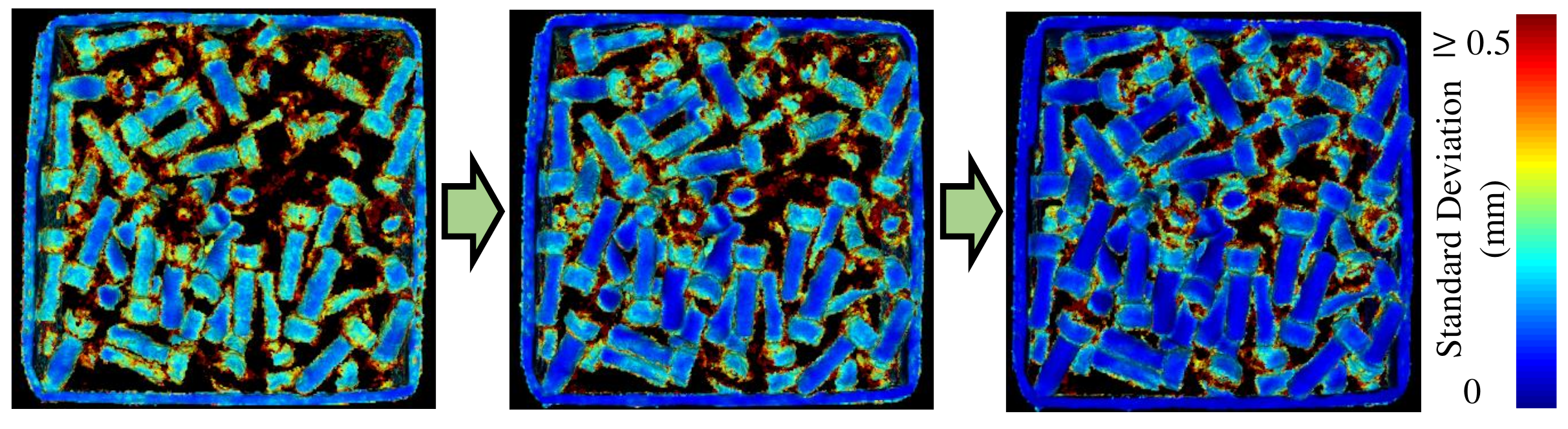}
  \vspace{-1.6\baselineskip}
  \caption{The propagation of standard deviation for inlier meshes.}
\end{subfigure}
\begin{subfigure}{0.45\textwidth}
  \includegraphics[width=\linewidth]{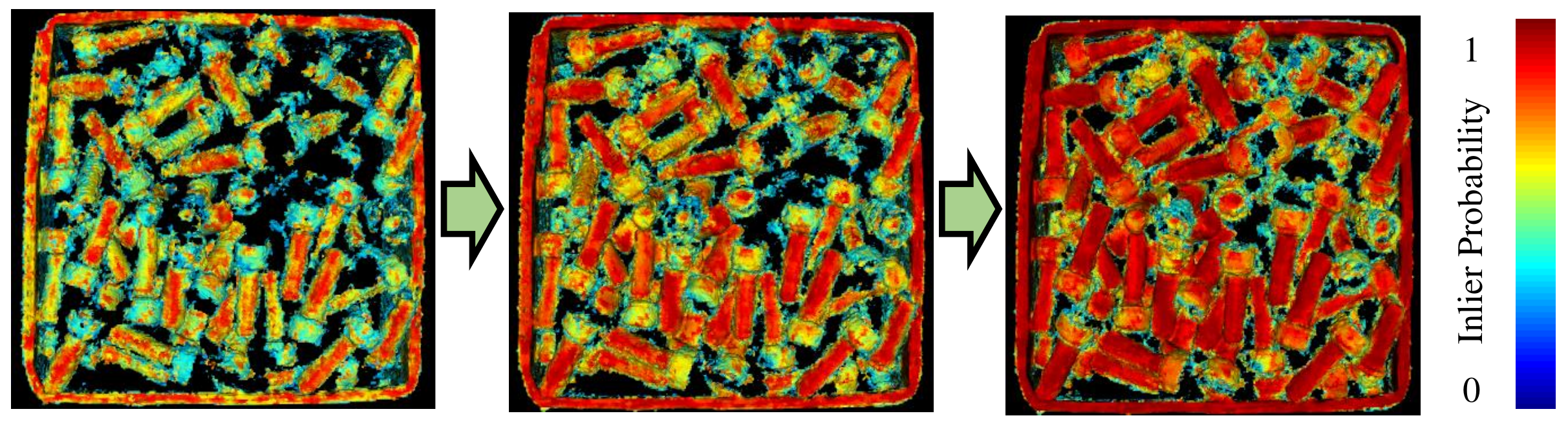}  
  \vspace{-1.6\baselineskip}
  \caption{The propagation of inlier probability.}
\end{subfigure}
\caption{Incremental scene updating for object "Chrome Screw". From left to right: Reconstruction from 10 views, 30 views and 88 views. The paired monochrome image is shown on the left of Figure \ref{fig7b}.}
\label{fig10}
\end{figure}

\subsection{Ground Truth Acquisition}
\textbf{Depth Maps.} Depth measurement by Ensenso camera suffers from large errors up to $2-3\,mm$ on reflective surfaces \cite{hodgson2017novel}, which are common to industrial parts. To acquire ground truth depths, we apply a scanning spray \cite{AESUB} on objects to create Lambertian surfaces, so that the Ensenso camera can achieve its optimal accuracy (less than $0.2\,mm$). The scanning spray generates a thin and homogeneous layer with $8-15\,\mu m$ thickness, which is one to two orders of magnitude less than the expected depth accuracy. As illustrated in Figure \ref{fig6}, we capture the ground truth and test images with two scans. In the initial scan, we apply the scanning spray and capture ground truth depth maps. After spray evaporation, the test images are captured during the second scan.

\textbf{Scene Model.} For each scene, we construct the ground truth mesh by applying TSDF fusion {\cite{newcombe2011kinectfusion}} on ground truth depth maps. To demonstrate that the reconstruction of the ground truth scene is not biased to any fusion method, we apply both the TSDF and our method to produce two sets of ground truth meshes, and compute mean point-to-point distance between these two meshes. The mean point-to-point distance is $0.03\,mm$, indicating our ground truth accuracy is not sensitive to different fusion methods.

\textbf{6D Object Poses.} To annotate 6D object poses of a bin instance, we first manually align object CAD models to ground truth scene model. To increase accuracy, we upsample the CAD model to a higher resolution and use ICP algorithm to refine 6D object poses. The misalignment can be identified from scene models, and poses were then manually adjusted. We repeat this process several times until a satisfactory alignment was achieved.

\section{EXPERIMENTS}
\begin{figure}[t]
\centering
\begin{subfigure}{0.45\textwidth}
    \includegraphics[width=0.32\linewidth]{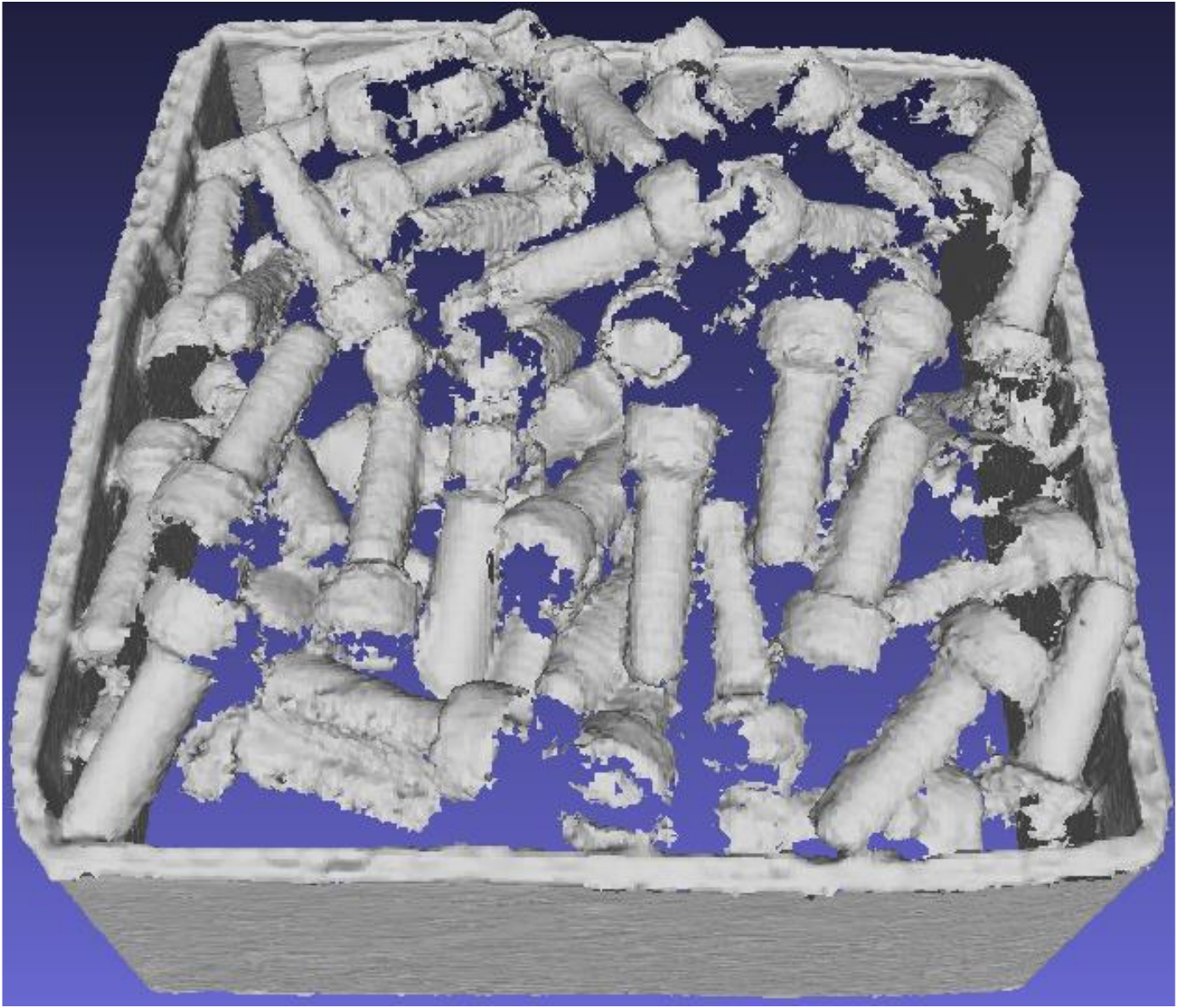}
    \includegraphics[width=0.32\linewidth]{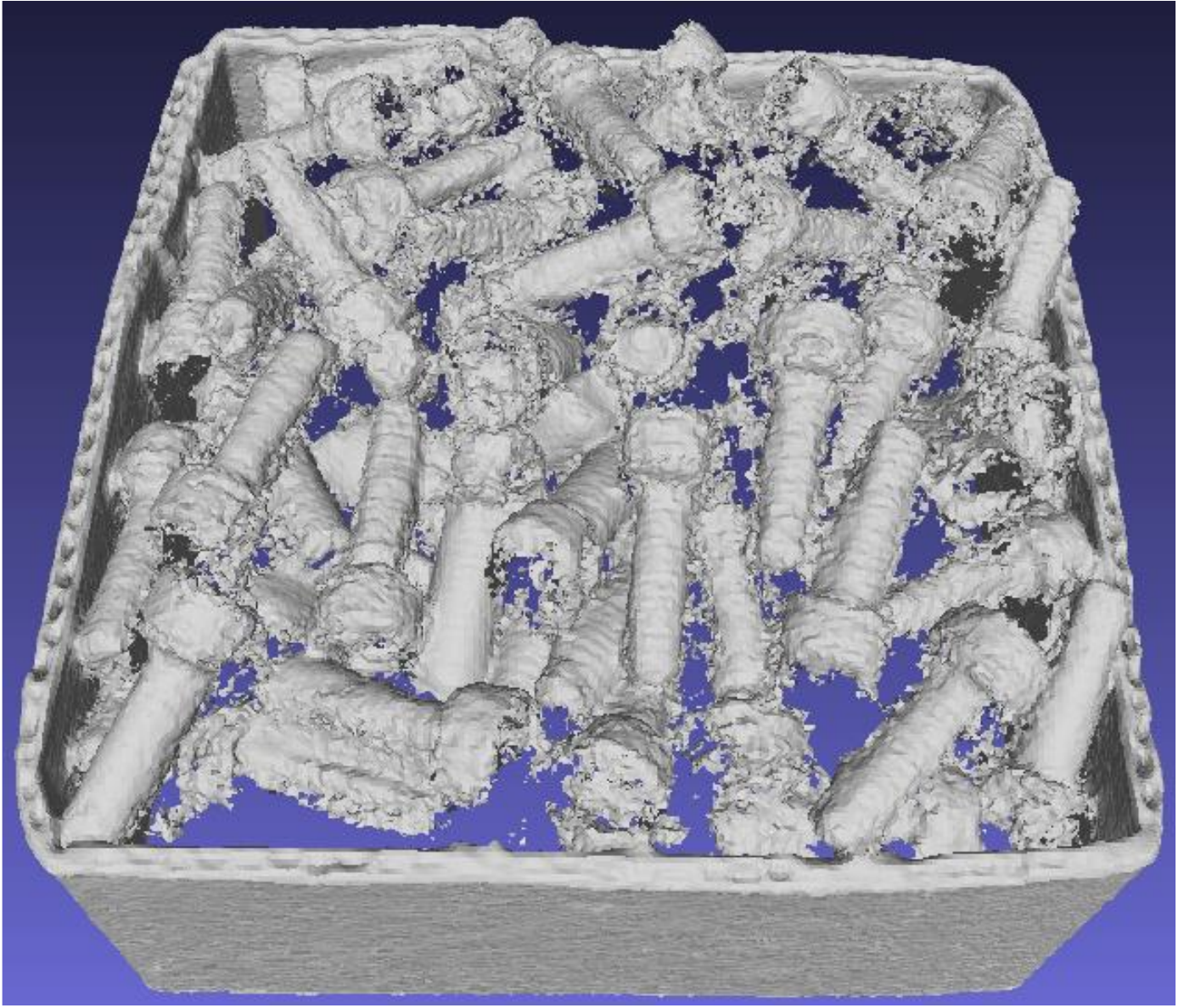}
    \includegraphics[width=0.32\linewidth]{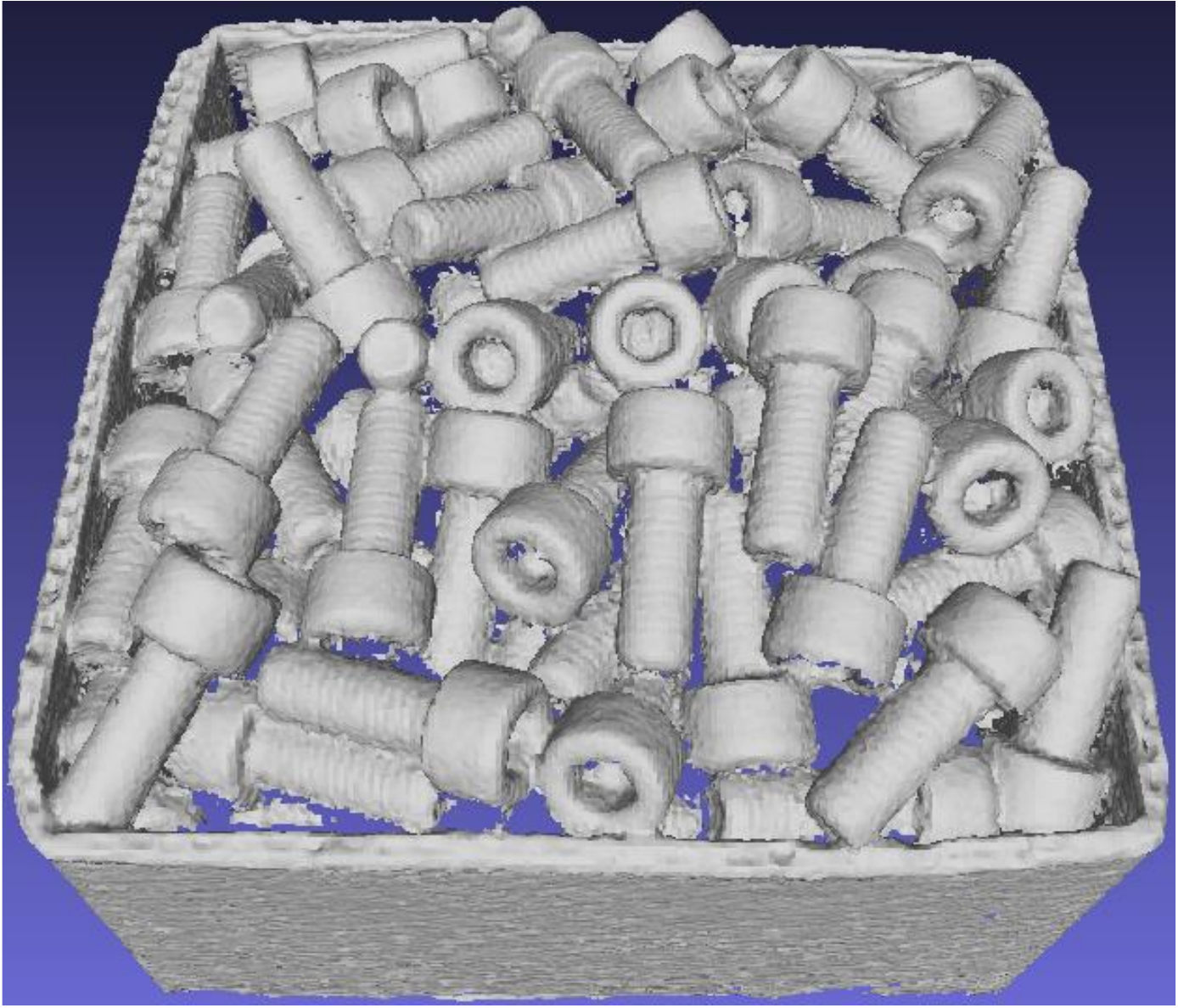}
    \vspace{-0.3\baselineskip}
    \caption{From left to right: Ours, TSDF, GT mesh.}
\label{fig9a}    
\end{subfigure}
\begin{subfigure}{0.45\textwidth}
\centering
    \includegraphics[width=0.45\linewidth]{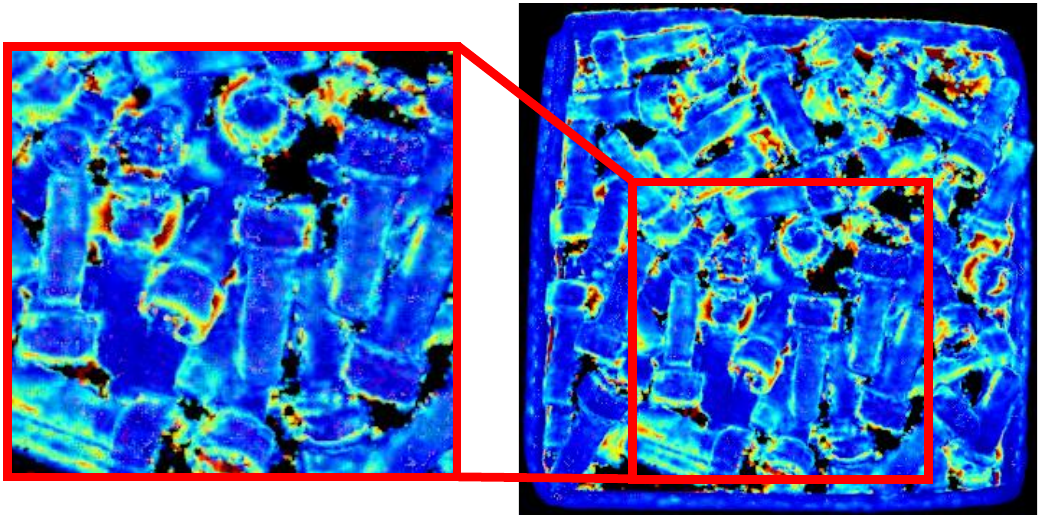}
    \includegraphics[width=0.52\linewidth]{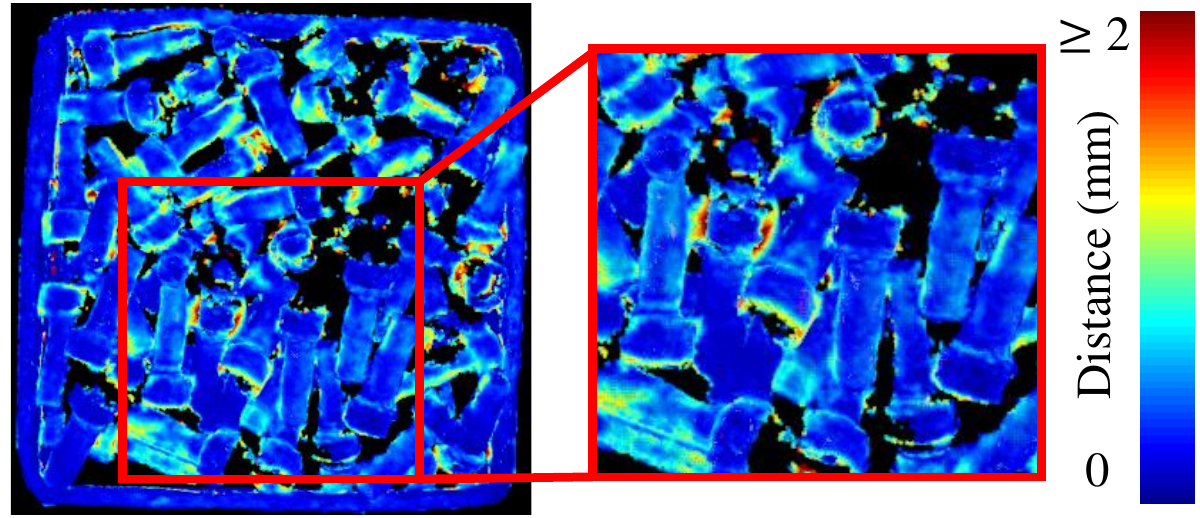}
    \vspace{-0.3\baselineskip}
    \caption{Left: TSDF. Right: Ours.}
\end{subfigure}
\caption{(a) Comparison of output mesh between TSDF and our framework for object "Chrome Screw" in a full-bin scene. (b) Heatmap of the point-to-point distance from reconstructed model to the ground truth mesh.}
\label{fig9}
\end{figure}

\begin{table*}[]
\resizebox{\textwidth}{!}{
\begin{tabular}{|c|c|c|c|c|c|c|c|c|c|c|}
\hline
\multirow{2}{*}{Bin} & \multirow{2}{*}{Object Category} & \multicolumn{3}{c|}{Mean Point-to-Point Distance (mm)} & \multicolumn{3}{c|}{Outlier Percentage (\%)} & \multicolumn{3}{c|}{Scene Completeness (\%)} \\ \cline{3-11} 
 &  & TSDF & TSDF w/ $W_{thr}$ & Ours & TSDF & TSDF w/ $W_{thr}$ & Ours & TSDF & TSDF w/ $W_{thr}$ & Ours \\ \hline
\multirow{3}{*}{Full} & Large Size & 0.42 & 0.29 & \textbf{0.25} & 18.3 & \textbf{3.18} & 4.1 & \textbf{99.5} & 98.8 & 99.3 \\ \cline{2-11} 
 & Complex Shape & 0.58 & 0.46 & \textbf{0.39} & 13.4 & 2.18 & \textbf{1.86} & \textbf{98.9} & 95.8 & 96.8 \\ \cline{2-11} 
 & High Gloss & 0.61 & 0.47 & \textbf{0.41} & 11.1 & 1.60 & \textbf{1.51} & \textbf{95.9} & 89.3 & 91.5 \\ \hline
\multirow{3}{*}{Low} & Large Size & 0.24 & 0.17 & \textbf{0.13} & 3.1 & \textbf{0.29} & 0.41 & \textbf{99.9} & 99.4 & 99.7 \\ \cline{2-11} 
 & Complex Shape & 0.44 & 0.34 & \textbf{0.3} & 3.28 & \textbf{0.35} & 0.38 & \textbf{90.1} & 78.5 & 79.9 \\ \cline{2-11} 
 & High Gloss & 0.43 & 0.36 & \textbf{0.24} & 1.88 & \textbf{0.16} & \textbf{0.16} & \textbf{85.03} & 74.5 & 77.0 \\ \hline
\multicolumn{2}{|c|}{Total} & 0.52 & 0.39 & \textbf{0.34} & 10.1 & \textbf{1.56} & 1.57 & \textbf{95.1} & 89.5 & 91.1 \\ \hline
\end{tabular}}
\caption{Quantitative reconstruction results in different object categories with varied bin scenarios.}
 \vspace{-0.8\baselineskip}
\label{tab1}
\end{table*}

The experiments regarding reconstruction and 6D pose estimation were performed on ROBI dataset, described in Section \ref{sec4}. Our framework is compared to traditional TSDF fusion as the baseline. To make the comparison fair, we also utilized a strategy from \cite{duhautbout2019distributed, fehr2017tsdf} to reduce outliers for TSDF. Specifically, all voxels that have received the number of measurements under a given threshold $W_{thr}$ are deleted. In our evaluation, parameter $W_{thr}$ is empirically set to $3$. For all experiments, we use a voxel size of $0.5\,mm$, the truncation distance is set to $1.5\,mm$.

\subsection{Reconstruction Evaluation}
We first visualize the propagation of surface uncertainties by rendering the inlier probability $\pi_k$ and standard deviation $\sigma_k$ in colormaps. As shown in Figure {\ref{fig10}}, when compared to the matte surface (bin wall), depth measurements for high gloss surface (objects in the bin) are more prone to noise and missed detections, hence high gloss reconstruction requires more views to converge. After the outlier rejection, the qualitative comparison, shown in Figure {\ref{fig9}}, illustrates that our framework produces smoother surfaces and fewer outliers than traditional TSDF.

For quantitative evaluation, we use three metrics by computing point-to-point distances between reconstructed mesh $\hat{\mathbf{M}}$ and ground-truth mesh $\mathbf{M}$. For each point in source mesh, its corresponding point is the closest point in target mesh. A point is defined as inlier if point-to-point Euclidean distance is smaller than $2\,mm$:
\begin{itemize}
\item \textbf{Mean Point-to-Point Distance.} We build correspondences from $\hat{\mathbf{M}}$ to $\mathbf{M}$, the mean point-to-point distance is computed over all inliers, and measures the reconstruction accuracy.
\item \textbf{Outlier Percentage.} Given correspondences from $\hat{\mathbf{M}}$ to $\mathbf{M}$, the outlier percentage is computed as the fraction of the number of outliers over the total number of vertices in the ground truth mesh.
\item \textbf{Scene Completeness.} We build correspondences from $\mathbf{M}$ to $\hat{\mathbf{M}}$, the scene completeness is computed as the fraction of the number of inliers over the total number of ground truth vertices.
\end{itemize}
These metrics are measured only on the surface of objects, which has a significant impact on later 6D object pose estimation. To illustrate reconstruction quality on different properties, we split all objects into three categories, based on their major characteristics: (a) Large Size ("Zigzag"), (b) Complex Shape ("DIN Connector" and "D-Sub Connector"), (c) High Gloss (remaining four objects). 

\begin{figure}[t]
\centering
\begin{subfigure}{.24\textwidth}
  \includegraphics[width=\linewidth]{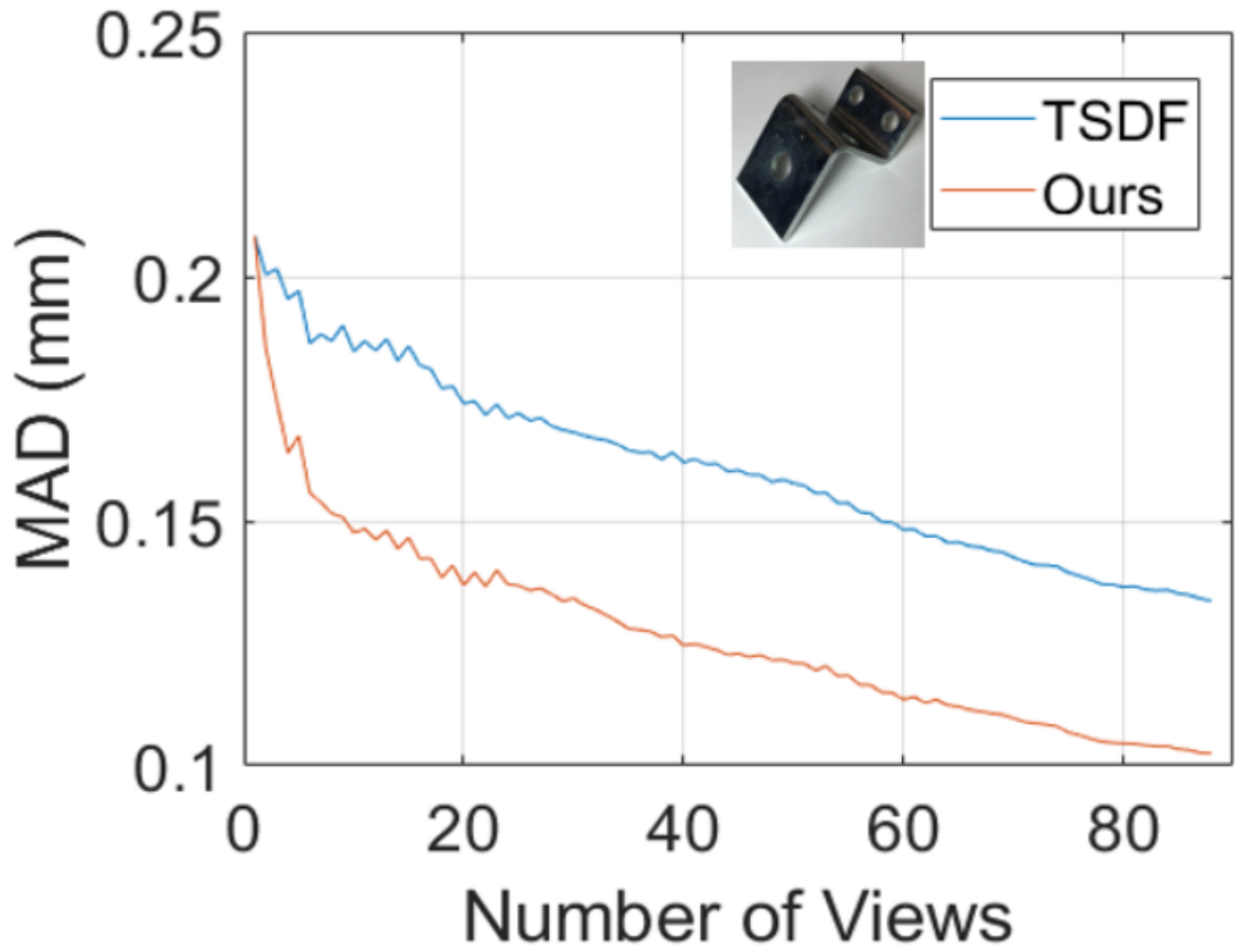}
\end{subfigure}
\begin{subfigure}{.24\textwidth}
  \includegraphics[width=\linewidth]{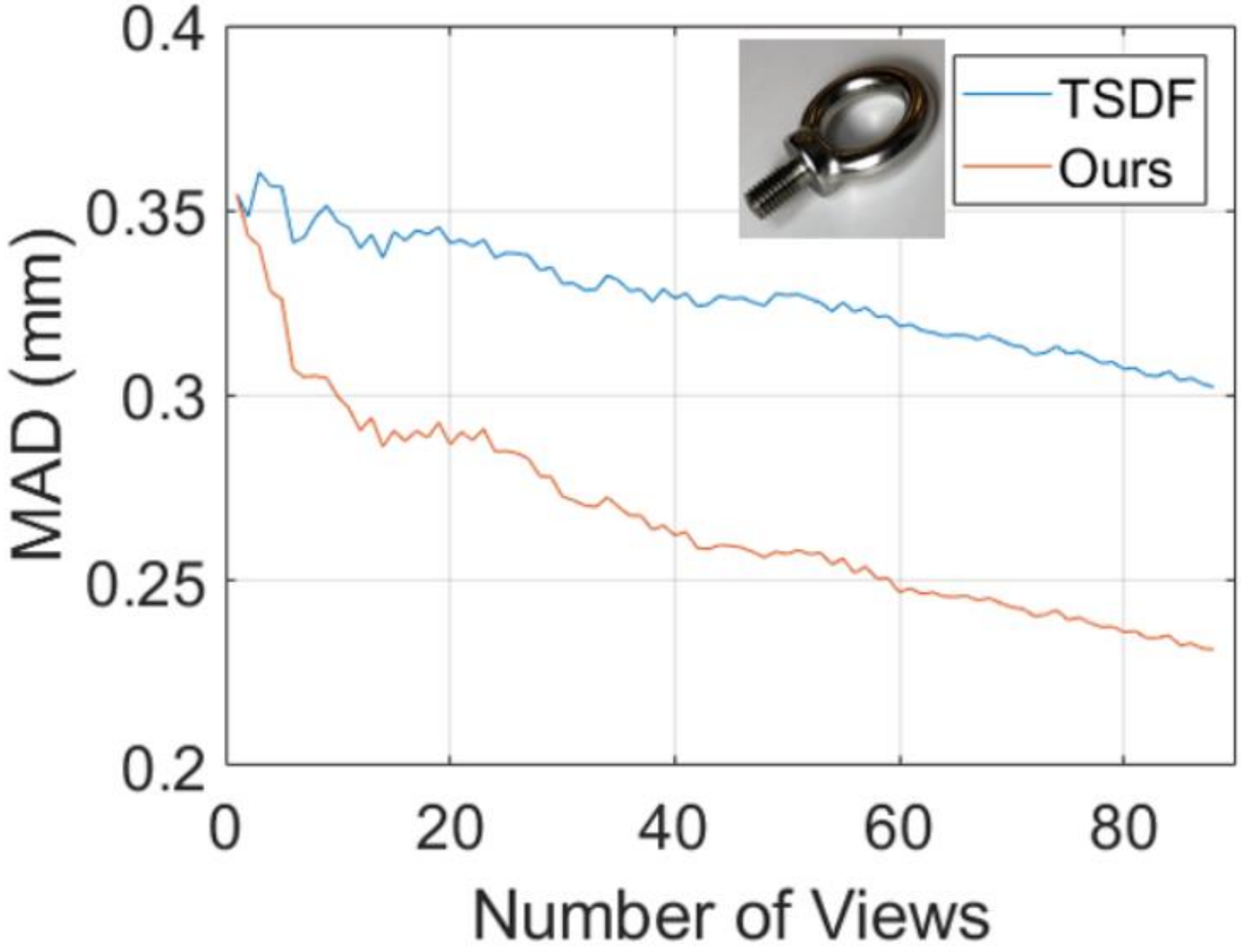}  
\end{subfigure}
\caption{Comparison of mean absolute distance (MAD) between TSDF and our methods. Left: Object "Zigzag" in full-bin. Right: Object "Eye Bolt" in full-bin.}
\label{fig8}
\end{figure}

Table \ref{tab1} shows the quantitative reconstruction results of each object category with the accumulation of all viewpoints. Our method not only achieves higher reconstruction accuracy (12.8\% improvement), but also makes a better trade-off between scene completeness and outlier percentage. Table {\ref{tab1}} also reveals that the reconstruction is particularly difficult for high gloss objects using both TSDF and our methods. This indicates that our method is upper bounded by depth measurement quality and the viewpoint coverage. Moreover, despite higher accuracy and fewer outliers, our method also sacrifices some completeness of the scene. This phenomenon is more obvious for low bin data, which have occlusions of bin walls and lower viewpoint coverage for object surfaces to converge.

To demonstrate the fast convergence of our method, we compute the mean absolute distance (MAD) of SDF value over all occupied voxels. As illustrated in Figure \ref{fig8}, as the accumulation of viewpoints, TSDF generally has a slow convergence in terms of MAD. In comparison, our method requires much fewer viewpoints to achieve same accuracy, making it highly useful for active vision techniques \cite{doumanoglou2016recovering}.

\begin{figure}[t]
\centering
\begin{subfigure}{.155\textwidth}
  \includegraphics[width=\linewidth]{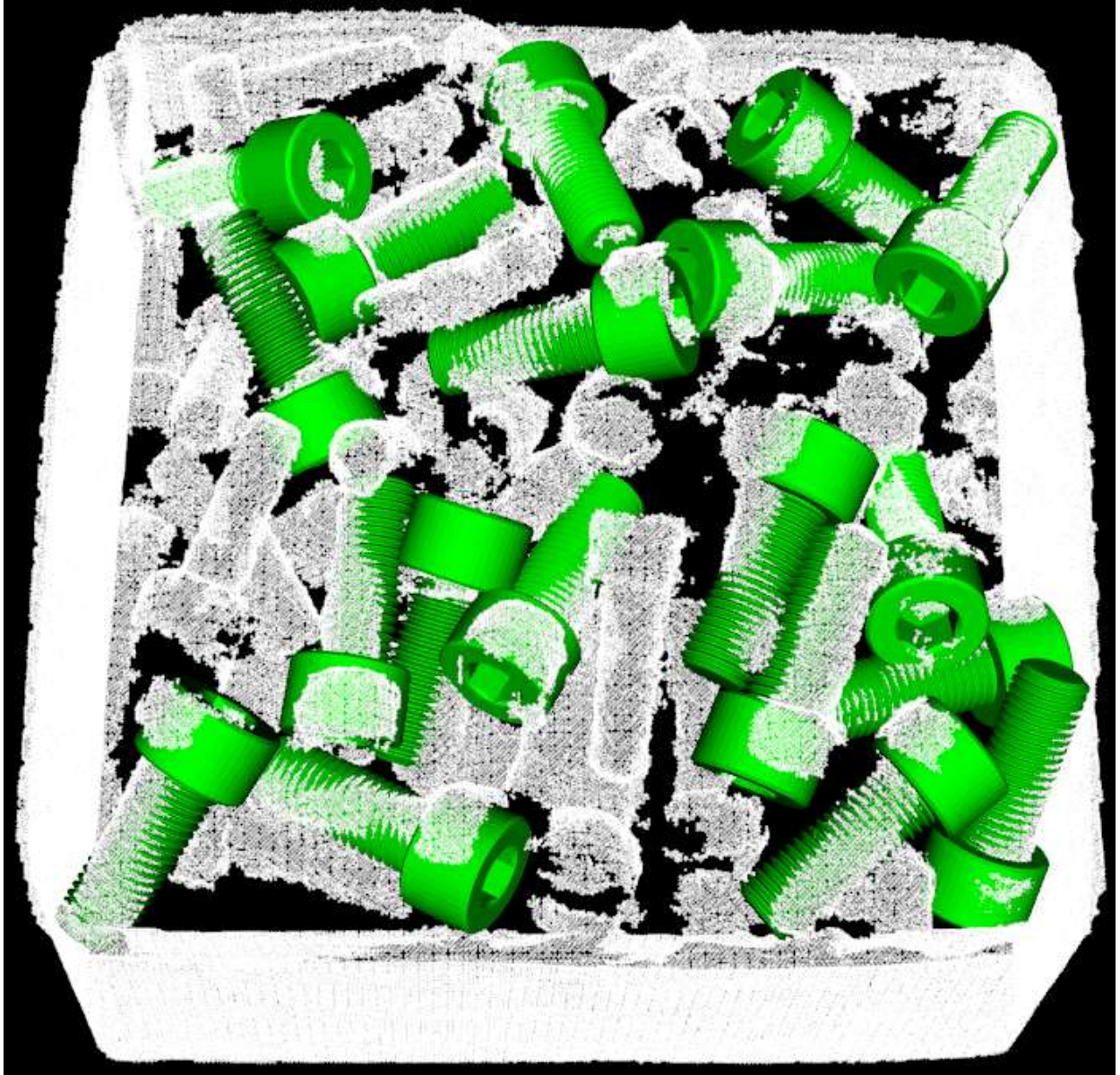}
  \caption{Ours}
\label{figrr2a}  
\end{subfigure}
\begin{subfigure}{.155\textwidth}
  \includegraphics[width=\linewidth]{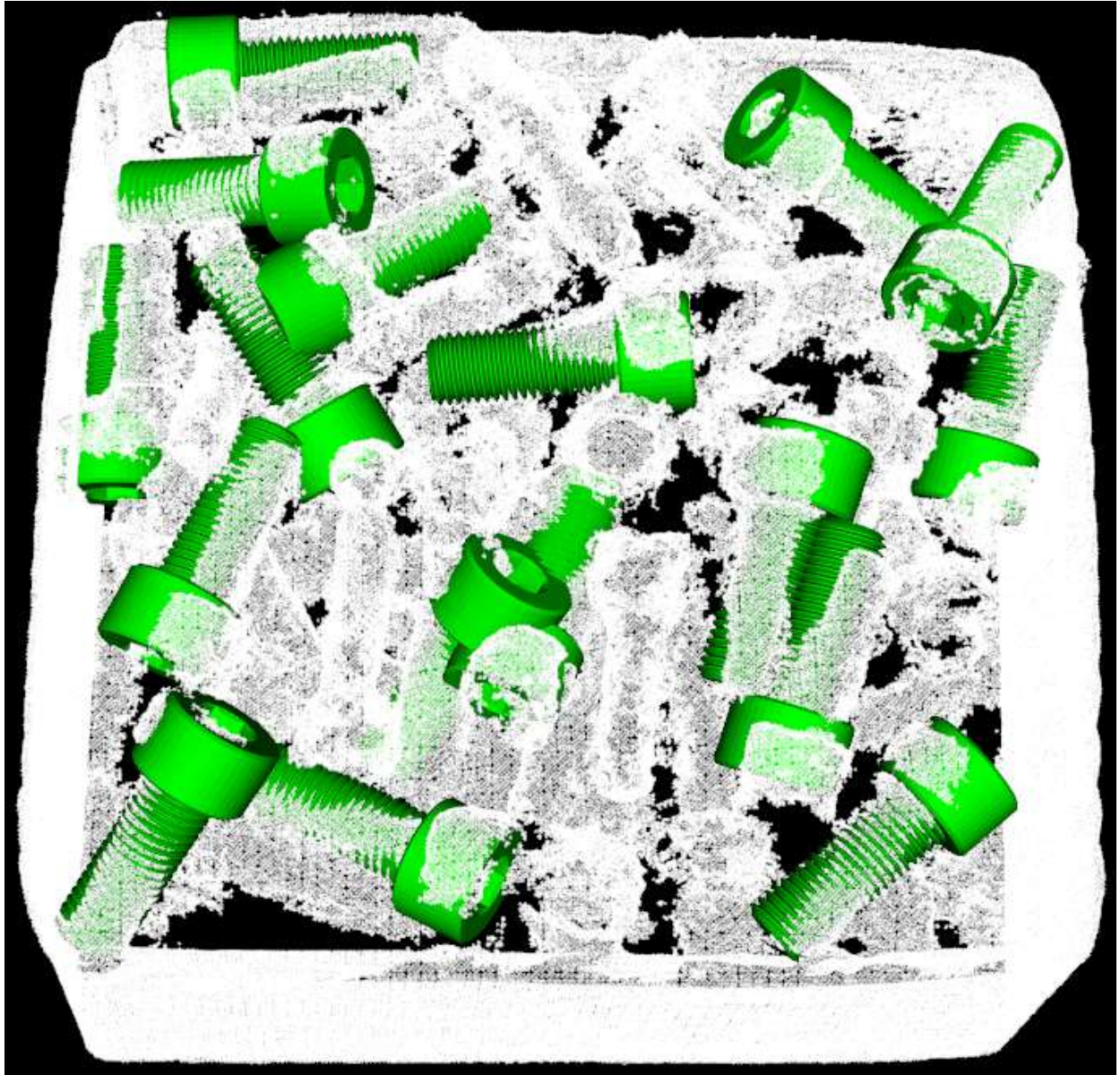}  
  \caption{TSDF}
\label{figrr2b}  
\end{subfigure}
\begin{subfigure}{.155\textwidth}
  \includegraphics[width=\linewidth]{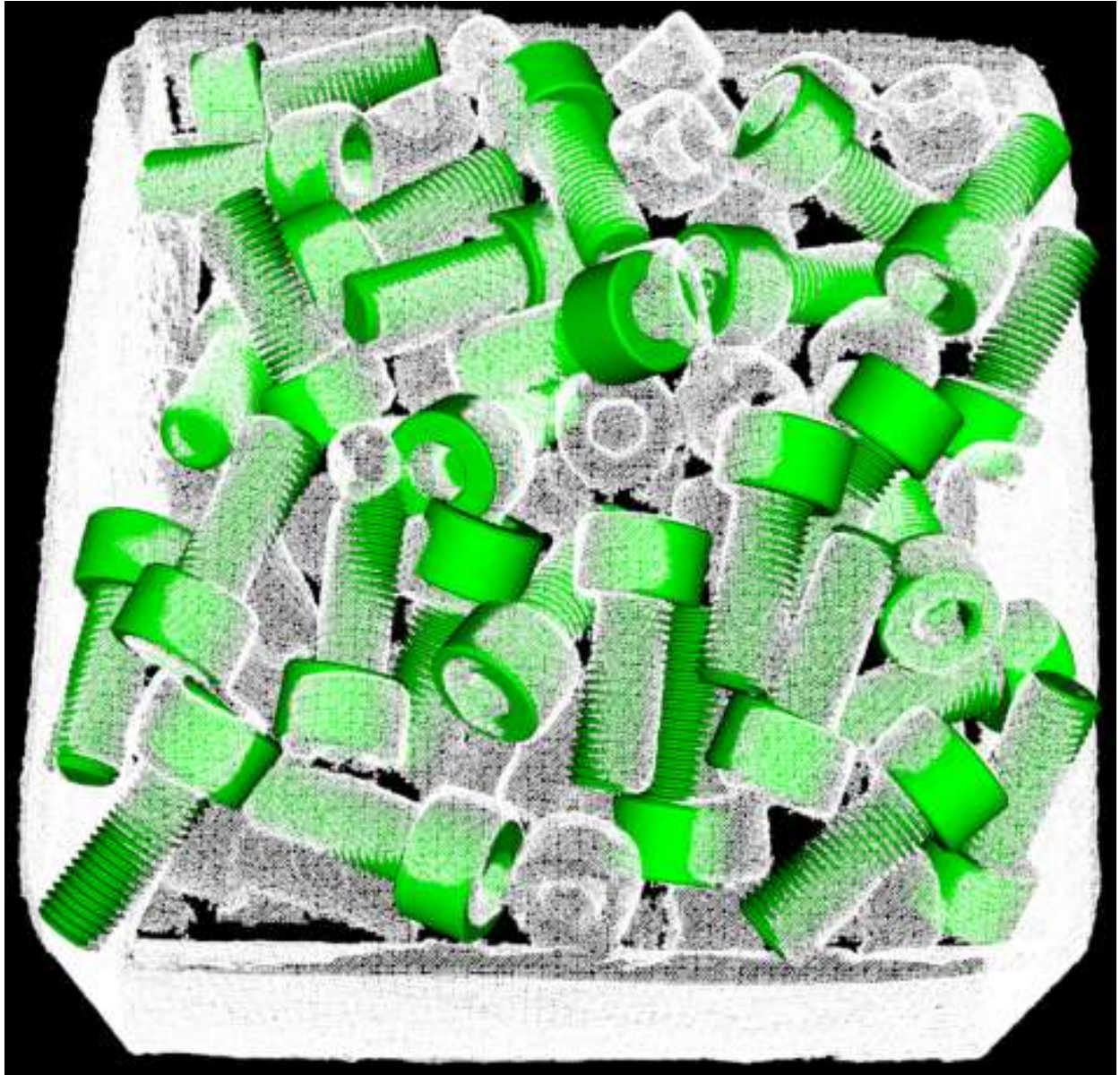}  
  \caption{GT mesh}
\label{figrr3b}  
\end{subfigure}
\caption{Correctly detected 6D poses for object "Chrome Screw" in a full-bin scene. The corresponding meshes are shown in Figure \ref{fig9}.}
\label{figrr2}
\end{figure}

\subsection{6D Object Pose Estimation Evaluation}
To evaluate the performance of 6D object pose estimation, we choose the algorithm proposed by Drost et al. \cite{drost2010model}. This method only relies on 3D point cloud data and solves object poses by coupling the idea of point-pair features (PPF) and a dense voting scheme. We compare our method against TSDF, its variant with thresholding strategy, ground truth mesh and the point cloud from a single (top center) viewpoint. All meshes are converted to point cloud by extracting their vertices.

As in \cite{drost2010model,rad2017bb8}, we compute the correct detection rate for each object and each scene. To measure the 6D pose error, we use the standard ADD score from \cite{hinterstoisser2012model}. A pose hypothesis is counted as correct when its ADD score is less than $10\%$ of the object diameter.

We present the correct detection rate in Table \ref{tab2}. It can be seen that, given optimal 3D data (GT mesh), the PPF pose estimator can provide close to perfect detection rates. In comparison, due to a large amount of missing depths, the detection performance is generally poor on single view based point cloud. This performance can be improved with multi-view volumetric depth fusion. And as a result of the complete scene and few outliers, our method outperforms TSDF and its variant by a large margin of 6.1\%. It is noteworthy that our improvements in detection rate are more significant for full-bin scenarios, which has more noise and outliers due to the clutter. Further comparison in Figure \ref{figrr2} illustrates that, as a result of the smoother surface and fewer outliers (shown in Figure \ref{fig9}), the PPF pose estimator provides more correct object poses on the mesh that was constructed by our method versus the TSDF method.

\begin{table}[]
\resizebox{0.485\textwidth}{!}{
\begin{tabular}{|c|c|l|c|c|c|c|c|}
\hline
\multirow{2}{*}{Bin} & \multicolumn{2}{c|}{\multirow{2}{*}{Object}} & \multicolumn{5}{c|}{Detection Rate (\%)} \\ \cline{4-8} 
 & \multicolumn{2}{c|}{} & \begin{tabular}[c]{@{}c@{}}Single\\ View\end{tabular} & \begin{tabular}[c]{@{}c@{}}TSDF\end{tabular} & \begin{tabular}[c]{@{}c@{}}TSDF\\w/ $W_{thr}$\end{tabular} & Ours & \begin{tabular}[c]{@{}c@{}}GT\\ Mesh\end{tabular} \\ \hline
\multirow{7}{*}{Full} & \multicolumn{2}{c|}{Zigzag} & 57.1 & 78.5 & 75.0 & \textbf{87.5} & 87.5 \\ \cline{2-8} 
 & \multicolumn{2}{c|}{Chrome Screw} & 0 & 33.3 & 37.2 & \textbf{49.0} & 83.3 \\ \cline{2-8} 
 & \multicolumn{2}{c|}{Gear} & 14.3 & 60.0 & 62.8 & \textbf{67.6} & 94.3 \\ \cline{2-8} 
 & \multicolumn{2}{c|}{Eye Bolt} & 42.1 & \textbf{78.9} & 68.4 & \textbf{78.9} & 97.3 \\ \cline{2-8} 
 & \multicolumn{2}{c|}{Tube Fitting} & 23.2 & 88.4 & 86.9 & \textbf{91.3} & 95.6 \\ \cline{2-8} 
 & \multicolumn{2}{c|}{DIN Connector} & 15.0 & 75.0 & 73.6 & \textbf{76.6} & 83.3 \\ \cline{2-8} 
 & \multicolumn{2}{c|}{D-Sub Connector} & 3.6 & 25.6 & 23.1 & \textbf{32.9} & 57.3 \\ \hline
\multicolumn{1}{|l|}{\multirow{7}{*}{Low}} & \multicolumn{2}{c|}{Zigzag} & 41.6 & \textbf{91.6} & \textbf{91.6} & \textbf{91.6} & 100 \\ \cline{2-8} 
\multicolumn{1}{|l|}{} & \multicolumn{2}{c|}{Chrome Screw} & 0 & 72.2 & 68.1 & \textbf{77.2} & 90.9 \\ \cline{2-8} 
\multicolumn{1}{|l|}{} & \multicolumn{2}{c|}{Gear} & 38.4 & 53.8 & 46.1 & \textbf{61.5} & 100 \\ \cline{2-8} 
\multicolumn{1}{|l|}{} & \multicolumn{2}{c|}{Eye Bolt} & 12.5 & 86.6 & 86.6 & \textbf{93.7} & 93.7 \\ \cline{2-8} 
\multicolumn{1}{|l|}{} & \multicolumn{2}{c|}{Tube Fitting} & 9.5 & 71.4 & 66.7 & \textbf{76.2} & 100 \\ \cline{2-8} 
\multicolumn{1}{|l|}{} & \multicolumn{2}{c|}{DIN Connector} & 9.1 & \textbf{63.6} & 31.8 & 59.1 & 90.9 \\ \cline{2-8} 
\multicolumn{1}{|l|}{} & \multicolumn{2}{c|}{D-Sub Connector} & 0 & \textbf{22.7} & 9.1 & \textbf{22.7} & 95.4 \\ \hline
\multicolumn{3}{|c|}{Total} & 14.9 & 58.1 & 54.9 & \textbf{64.2} & 85.9 \\ \hline
\end{tabular}}
\caption{Detection rate results for different objects.}
\label{tab2}
\end{table}

\section{CONCLUSIONS}
In this paper, we proposed a probabilistic framework for scene reconstruction in the bin-picking problem. Based on active stereo camera data, we explicitly estimated two types of depth uncertainties, and incorporated them into a probabilistic framework for incrementally updating of the scene. The high-quality mesh can be used as the input 3D data for improving the performance of 6D object pose estimation. To evaluate the performance of reconstruction and object pose estimation, we construct ROBI, a real-world dataset of reflective objects in bin-picking scenarios. Our approach outperforms the traditional TSDF for both reconstruction and 6D object pose estimation. As a future work, we want to investigate how to strategically select viewpoints, and achieve the optimal performance of reconstruction accuracy as well as 6D object pose estimation with the minimum number of views.

\addtolength{\textheight}{-12cm}   % This command serves to balance the column lengths
                                  % on the last page of the document manually. It shortens
                                  % the textheight of the last page by a suitable amount.
                                  % This command does not take effect until the next page
                                  % so it should come on the page before the last. Make
                                  % sure that you do not shorten the textheight too much.

%%%%%%%%%%%%%%%%%%%%%%%%%%%%%%%%%%%%%%%%%%%%%%%%%%%%%%%%%%%%%%%%%%%%%%%%%%%%%%%%

%%%%%%%%%%%%%%%%%%%%%%%%%%%%%%%%%%%%%%%%%%%%%%%%%%%%%%%%%%%%%%%%%%%%%%%%%%%%%%%%

%%%%%%%%%%%%%%%%%%%%%%%%%%%%%%%%%%%%%%%%%%%%%%%%%%%%%%%%%%%%%%%%%%%%%%%%%%%%%%%%

%%%%%%%%%%%%%%%%%%%%%%%%%%%%%%%%%%%%%%%%%%%%%%%%%%%%%%%%%%%%%%%%%%%%%%%%%%%%%%%%

\bibliographystyle{ieeetr}
\bibliography{output.bbl}

\end{document}